\documentclass[letterpaper, 10 pt, conference]{ieeeconf}  

\usepackage{cite}
\usepackage[dvips]{graphicx}
\usepackage[cmex10]{amsmath}
\usepackage{color}
                      
\usepackage{subfloat}
\usepackage{subfig}
\usepackage{overpic}
\usepackage{array}                      

\IEEEoverridecommandlockouts                              

\title{\LARGE \bf
{A Fast Free-viewpoint Video Synthesis Algorithm for Sports Scenes}
}

\author{
Jun Chen, Ryosuke Watanabe, Keisuke Nonaka, Tomoaki Konno, Hiroshi Sankoh, and Sei Naito
\thanks{
J. Chen, R. Watanabe, K. Nonaka, T. Konno, H. Sankoh, S. Naito
are with Ultra-realistic Communication Group, KDDI Research, Inc., Fujimino, Japan (corresponding author (J. Chen) Tel: +81-70-3825-9914; e-mail: ju-chen@kddi-research.jp).}}

\begin{document}

\maketitle
\thispagestyle{empty}
\pagestyle{empty}

\begin{abstract}

In this paper, we report on a parallel free-viewpoint video synthesis algorithm that can efficiently reconstruct a high-quality 3D scene representation {of} sports scenes.
The proposed method focuses on {a} scene that is captured by multiple synchronized cameras featuring wide-baselines.
The following strategies are introduced to accelerate the production of a free-viewpoint video {taking the improvement of visual quality into account}:
(1) a sparse point cloud is reconstructed using {a volumetric visual hull approach}, and an exact 3D ROI is found for each object using an efficient connected components labeling algorithm. {Next,} the reconstruction of a dense point cloud is accelerated by implementing visual hull only in the ROIs;
(2) an accurate polyhedral surface mesh is built by estimating {the} exact intersections between grid cells and {the} visual hull;
(3) the appearance of {the} reconstructed presentation is reproduced {in} a view-dependent manner that respectively renders the non-occluded and occluded region with the nearest camera and its neighboring cameras.
The production for {volleyball and judo} {sequences} demonstrates the effectiveness of our method in terms of {both} execution time and visual quality.

\end{abstract}

\section{INTRODUCTION}

{{F}}ree-viewpoint video (FVV) is {{a well-known technique that provides an}} immersive user experience {{when}} viewing visual media.  
Compared with traditional fixed-viewpoint video, it allows users to select a viewpoint interactively and is capable of rendering a new view from a novel viewpoint. 
Since a virtualized reality system \cite{kanade1997virtualized} that distributes $51$ cameras over a $5~m$ dome with controlled lighting and well-calibrated cameras was introduced, FVV has been a long-standing research topic in the field of computer vision ranging from model construction of a static object for films \cite{matsuyama2002generation} to the generation of dynamic object models for sports scenes \cite{guillemaut2011joint, germann2012novel, nonaka2018}.
Moreover, {this has not been confined to academia}, the companies LiberoVision, Intel, and 4DViews also attach importance to the technique {{and have been providing}} visual effects applications for various purposes.

Techniques for rendering a free-viewpoint video for sports scenes from multiple cameras in an uncontrolled environment can be categorized into two classes: billboard-based \cite{Sankoh2018Acmmm, hayashi2006synthesizing, nonaka2017billboard, nonaka2018optimal, Sabirin2018Toward} and model-based methods \cite{kilner2007dual, liu2010point, slabaugh2001survey, seitz1999photorealistic, esteban2004silhouette, sinha2005multi, starck2007surface}. 
Billboard-based methods construct a single planar billboard for each object, acquire the visual texture from the nearest camera, and estimate the 3D position of each object using geometric properties among {{the}} cameras \cite{Sankoh2018Acmmm} or {utilize a} deep learning based method \cite{rematas2018soccer}. 
The billboards rotate around a specific axis with the movement of the viewpoint {providing} a walk-through and fly-through experience.
{It achieves good results with very little overhead incurred in the construction process.}
However, the transition between views is not smooth because the billboard is constructed only for the viewpoint where a camera is placed.
Another {issue to overcome with} these methods is occlusion, that is, multiple objects obscure each other in each camera.   
{M}odel-based methods describe a scene by {means of a} 3D mesh \cite{kilner2007dual} or point clouds \cite{liu2010point}.
The appearance of a scene is reproduced by mapping a corresponding texture onto the 3D model.
These methods offer the functionality of full freedom of virtual view and continuous change in appearance.
Visual {h}ull \cite{cheung2000real, smolic20113d} is a 3D reconstruction technique that approximates the 3D model of an object by back-projecting foreground silhouettes into 3D space.
With the advantage of low algorithmic complexity and robustness on calibration, it is often used in FVV production.
However, the computation time and memory consumption grow rapidly as the resolution of a pre-defined 3D volume {increases}.
The shape-from-photo-consistency \cite{slabaugh2001survey, seitz1999photorealistic} computes a more accurate 3D approximation of the scene, but it is usually sensitive to calibration errors and object textures.  
There are also some hybrid methods \cite{esteban2004silhouette, sinha2005multi, starck2007surface} that combine photo-consistency constraints, silhouette constraints, and sparse feature correspondence to reconstruct a scene with {a} high {degree of} accuracy.
However, the simultaneous process of several constraints makes them impossible to accomplish a production in a short time.

\begin{figure*}[t]
\centering
\footnotesize
	\begin{minipage}[b]{0.1375\linewidth}
 		\centering
 		\subfloat[]
		{
 	 		\begin{overpic}[width=1\textwidth]
 	 			{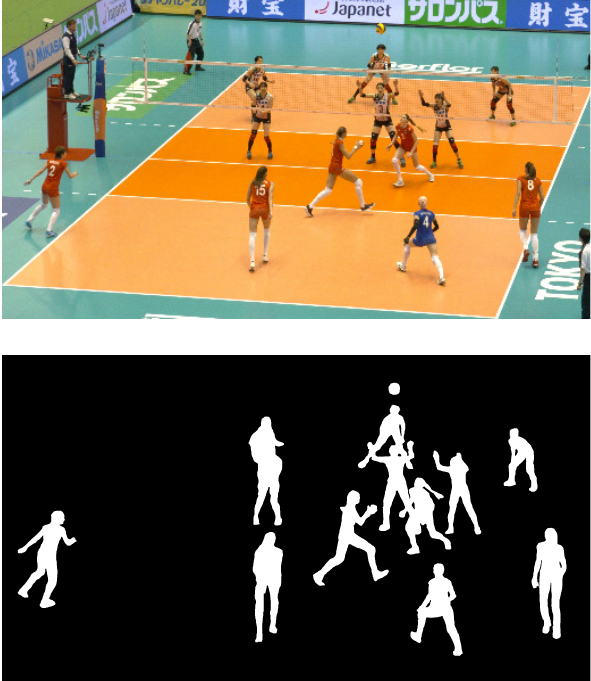}
 		\end{overpic}
 	 	}
	\end{minipage}
	\begin{minipage}[b]{0.22\linewidth}
 		\centering
 		\subfloat[]
		{
 			\begin{overpic}[width=1\textwidth]
 	 			{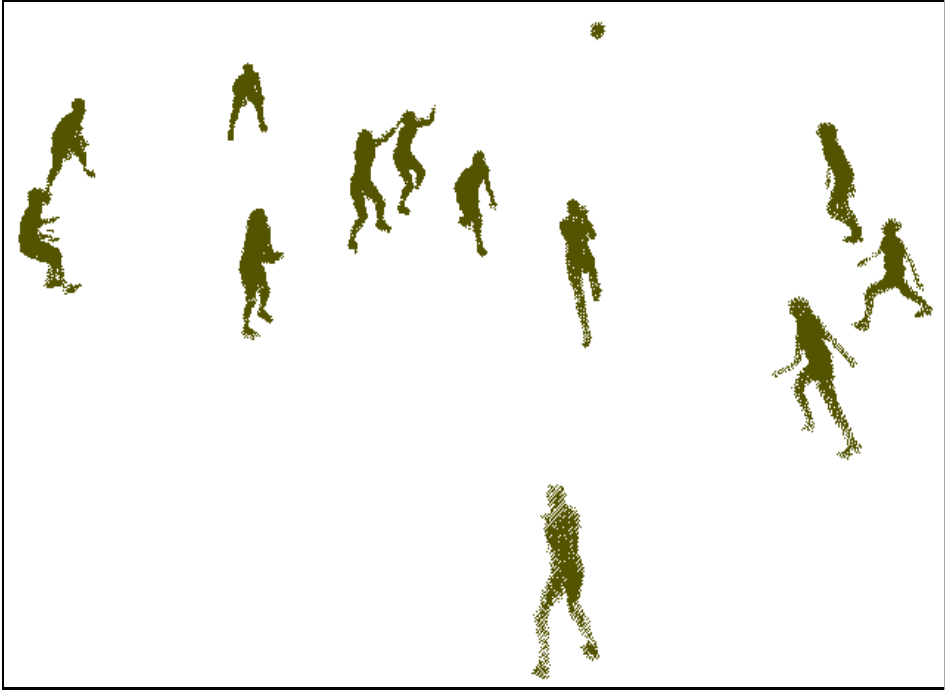}
 		\end{overpic}
 		}
	\end{minipage}
	\begin{minipage}[b]{0.22\linewidth}
 		\centering
 		\subfloat[]
		{
 			\begin{overpic}[width=1\textwidth]
 	 			{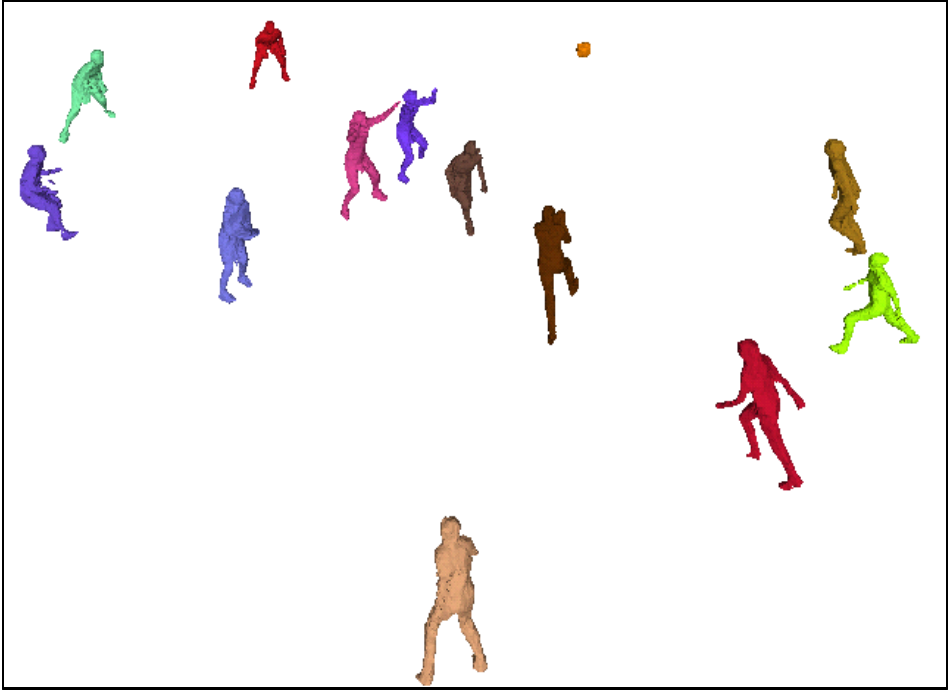}
 		\end{overpic}
 		}
	\end{minipage}
	\begin{minipage}[b]{0.1804\linewidth}
 		\centering
 		\subfloat[]
		{
 			\begin{overpic}[width=1\textwidth]
 	 			{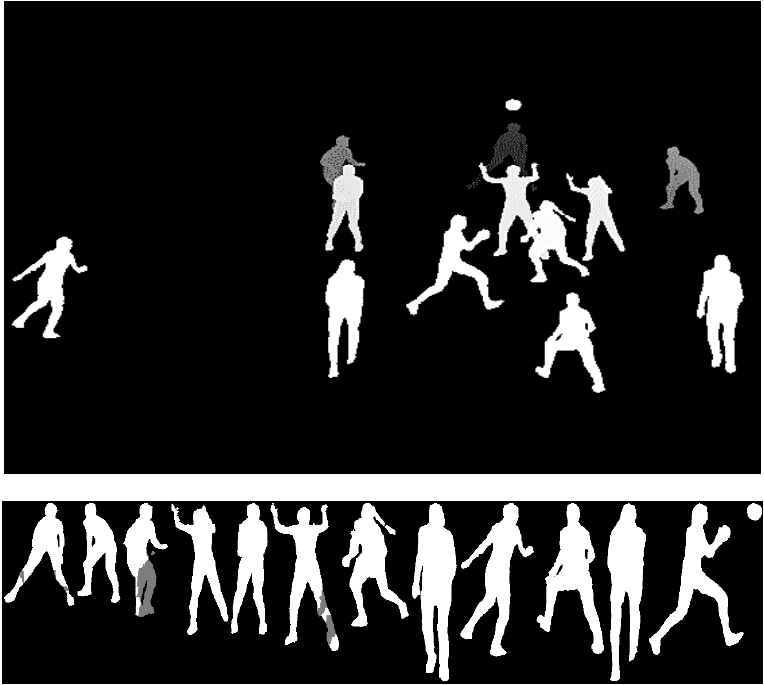}
 		\end{overpic}
 		}
	\end{minipage}		
	\begin{minipage}[b]{0.22\linewidth}
 		\centering
 		\subfloat[]
		{
 			\begin{overpic}[width=1\textwidth]
 	 			{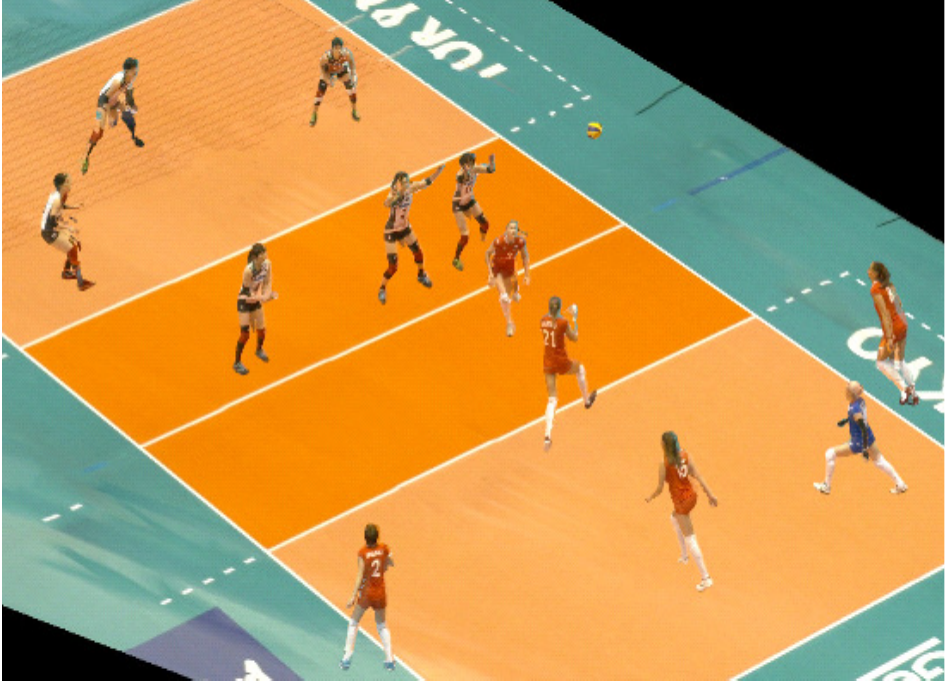}
 		\end{overpic}
 		}
	\end{minipage}		
\caption{Overview of our FVV synthesis method. (a) A color image and its silhouette. (b) Volumetric visual hull reconstruction. (c) Surface polygonization. (d) Visibility detection (e) View-dependent {r}endering.}
\label{fig:flowchart}
\vskip -4mm
\end{figure*}

{The goal of our research is to find a solution that reconstructs a 3D scene representation for sports event efficiently and improves the visual quality of synthesized virtual images.
To achieve it, we propose a GPU-based parallel FVV synthesis algorithm whose main contributions are: }
(1), a {coarse-to-fine} volumetric visual hull reconstruction is performed to reduce the computation time of the 3D shape approximation for a large space;
(2), an accurate polyhedral mesh is built by estimating the exact intersections between grid cells and the visual hull boundary, which smooths the mesh surface while retaining the robustness of the {visual hull};
(3), a view-dependent rendering method is performed to improve the visual quality of synthesized images, in which the nearest camera renders the non-occluded parts while its neighboring cameras render the occluded parts.
In the following sections, we will explain our method in detail, demonstrate its performance by comparing it to existing algorithms with {volleyball and judo} {{sequences}}, and discuss the influence of parameters on time complexity.


%
%
%
%

\section{{Proposed parallel algorithm}}

Fig.~\ref{fig:flowchart} represents the processing flow of our algorithm{.} 
{It} comprises four steps{:} volumetric visual hull reconstruction, surface polygonization, visibility detection, and view-dependent rendering.
{In addition to these}, some pre-processes such as camera calibration and silhouette extraction are carried out.

\subsection{Pre-processes}

In our setting, the cameras {remain} static while recording the {scene.} 
{This allows the cameras to be calibrated in advance using the camera model proposed by Jean-Yves Bouguet \cite{bouguet2004camera}.}
What should {be} note{d} here is that each camera is calibrated individually without the involvement of stereo camera calibration.

\begin{figure}[t]
\centering
\footnotesize
	\begin{minipage}[b]{0.48\linewidth}
 		\centering
 		\subfloat[]
		{
 	 		\begin{overpic}[width=1\textwidth]
 	 			{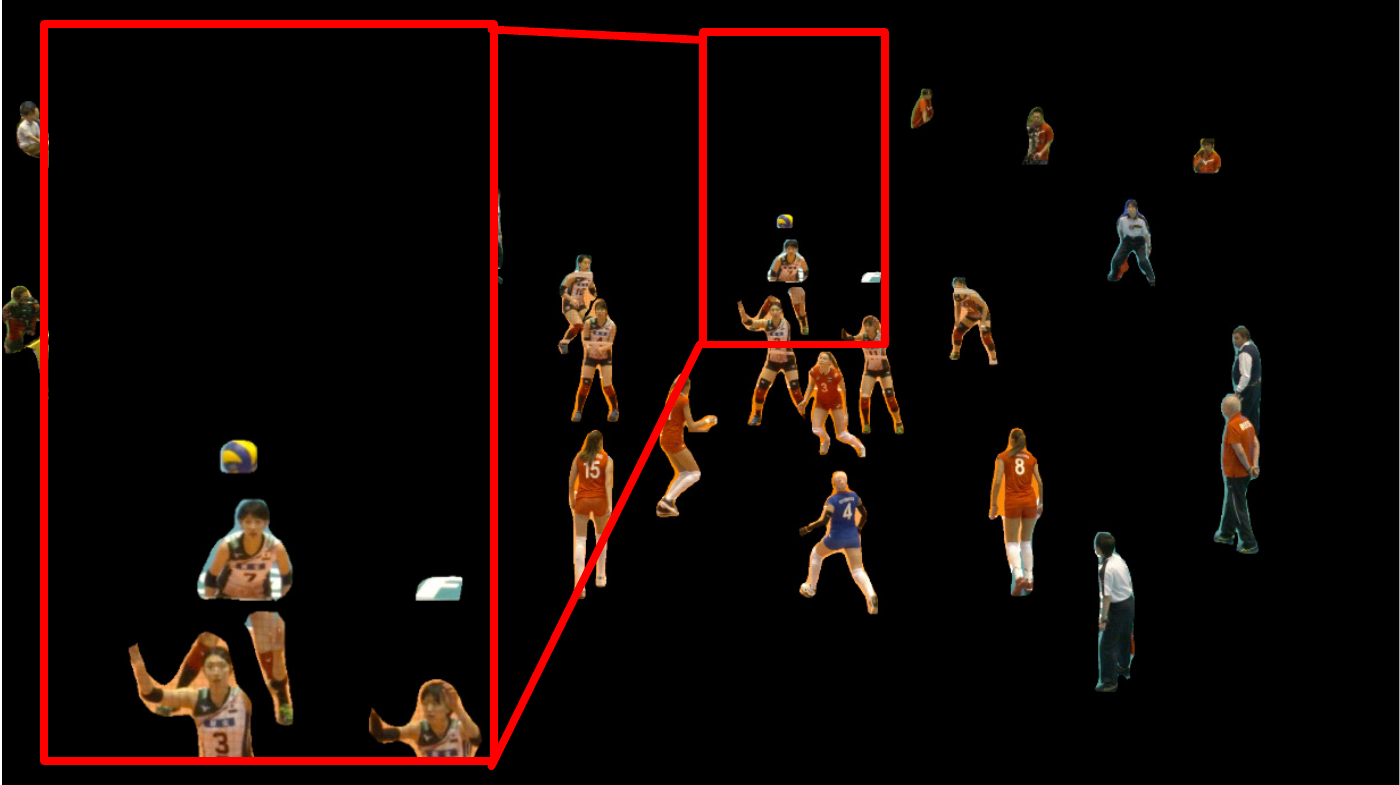}
 		\end{overpic}
 	 	}
	\end{minipage}
\hskip 1mm
	\begin{minipage}[b]{0.48\linewidth}
 		\centering
 		\subfloat[]
		{
 			\begin{overpic}[width=1\textwidth]
 	 			{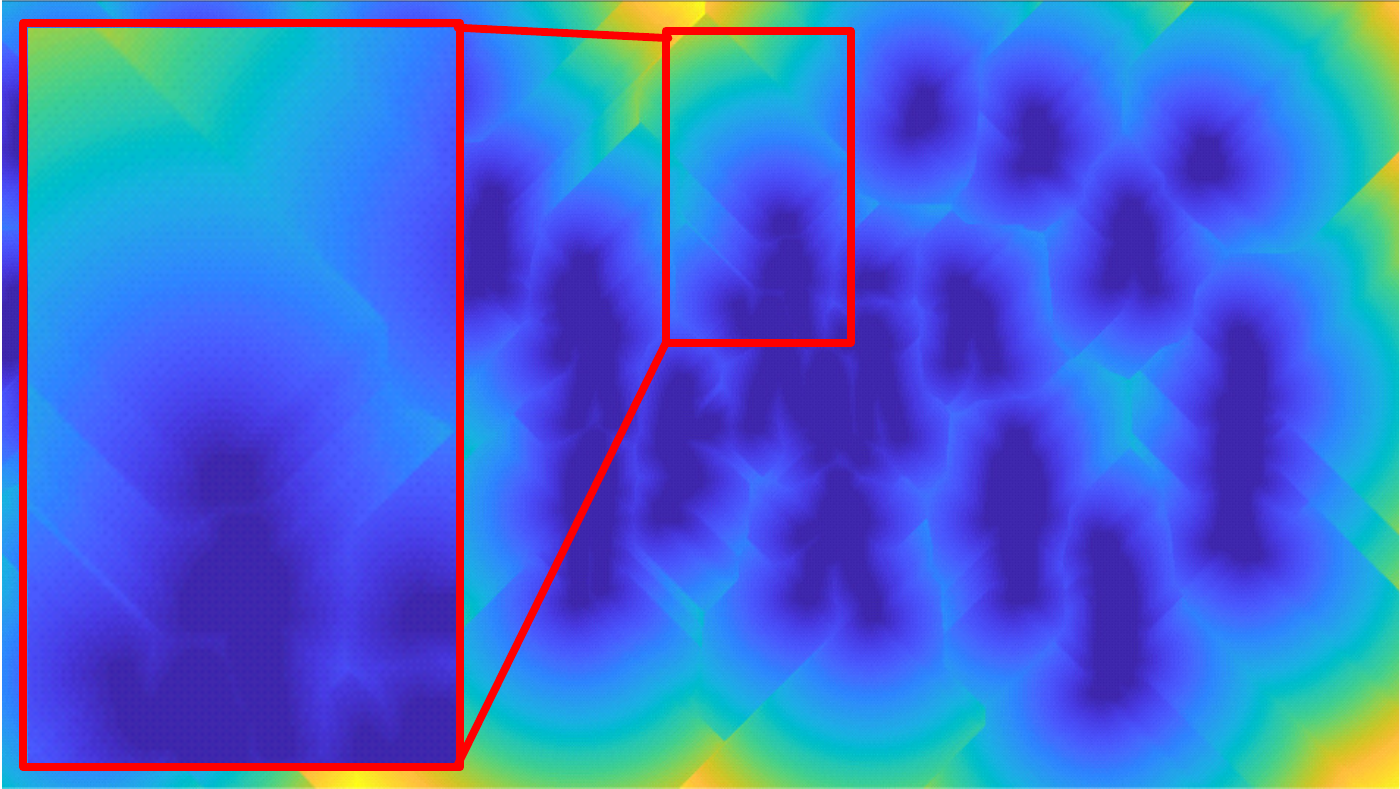}
 		\end{overpic}
 		}
	\end{minipage}
\vskip 1mm
	\begin{minipage}[b]{0.48\linewidth}
 		\centering
 		\subfloat[]
		{
 			\begin{overpic}[width=1\textwidth]
 	 			{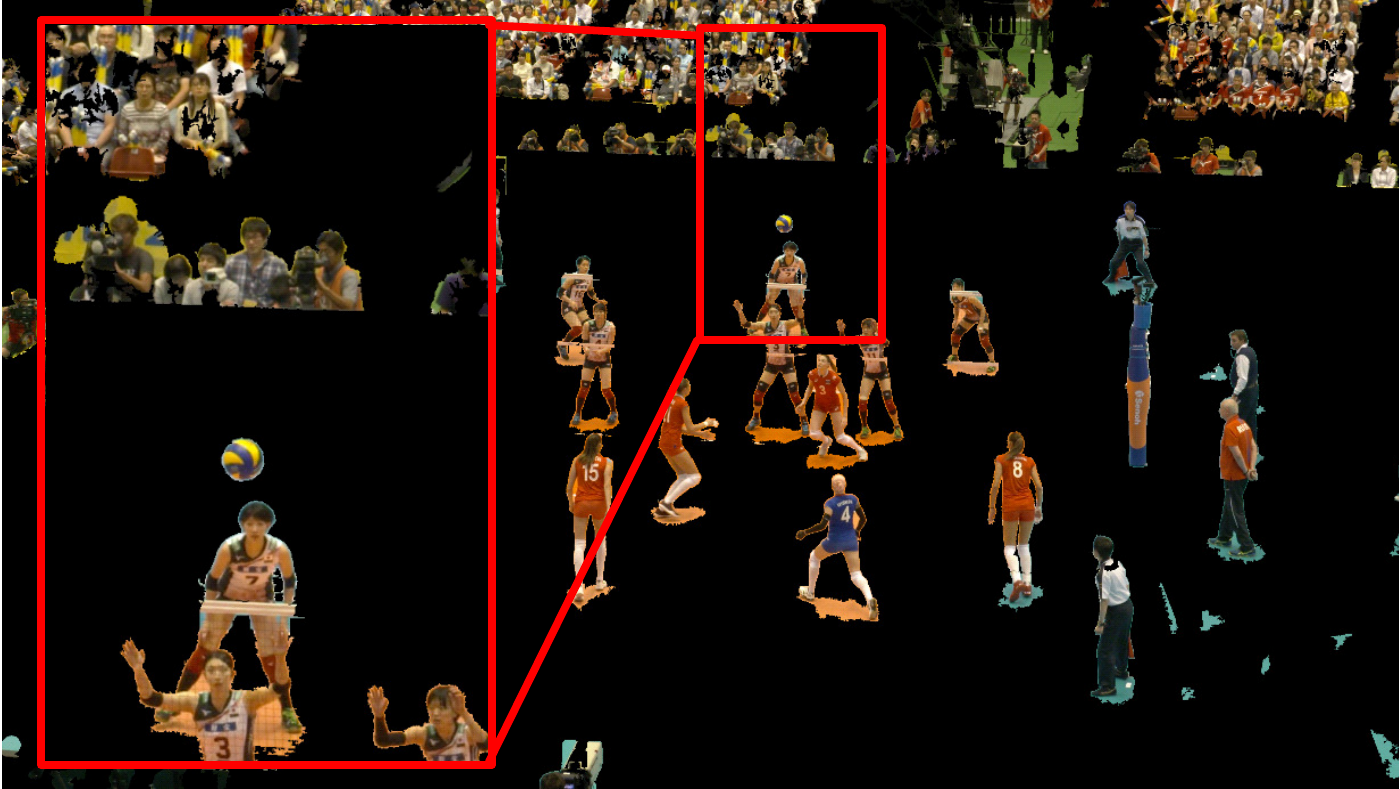}
 		\end{overpic}
 		}
	\end{minipage}
\hskip 1mm
	\begin{minipage}[b]{0.48\linewidth}
 		\centering
 		\subfloat[]
		{
 			\begin{overpic}[width=1\textwidth]
 	 			{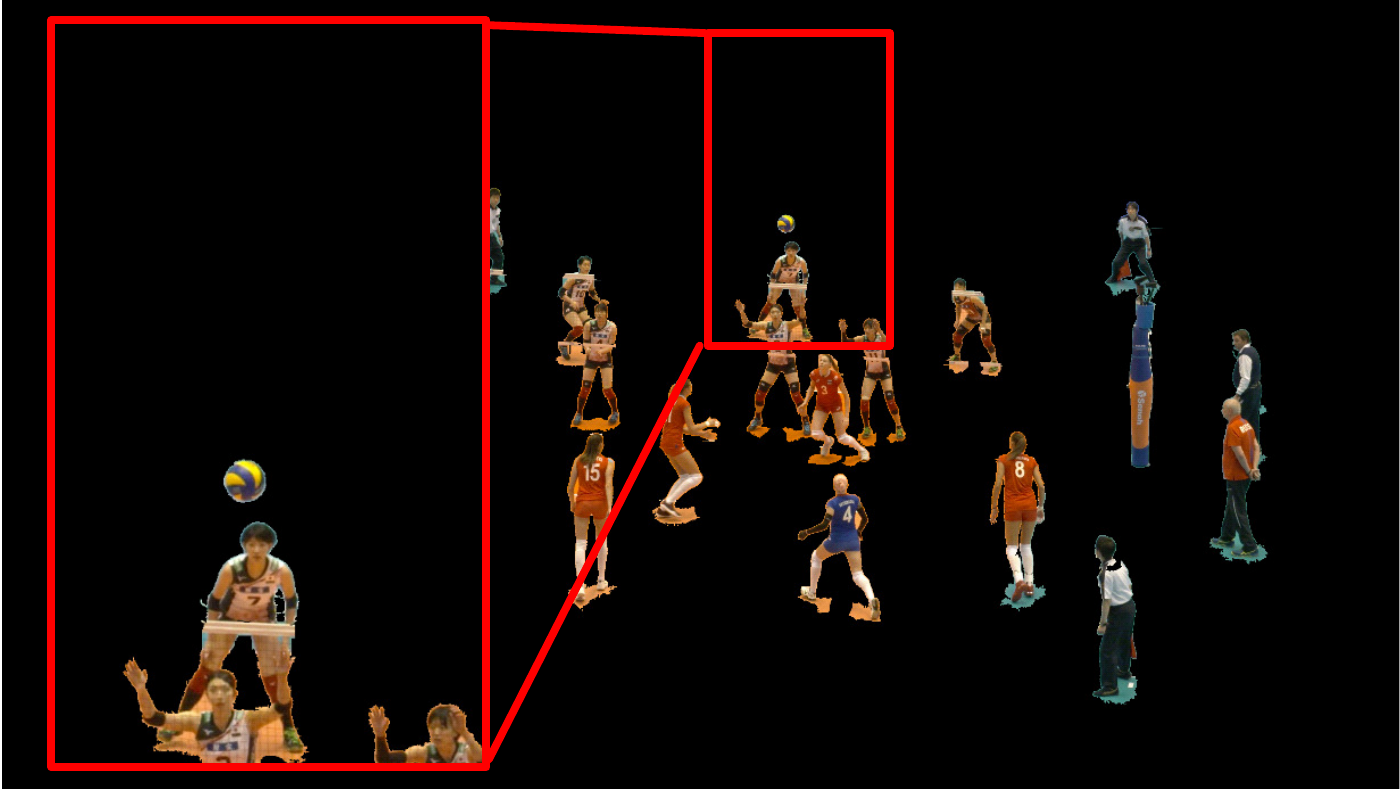}
 		\end{overpic}
 		}
	\end{minipage}			
\caption{Silhouette extraction. The red rectangular is the enlarged views of a selected region. {(a) Predicted foreground region by Mask-RCNN.} (b) Distance map. {(c) Segmentation by the fixed-parameter background modeling method.} (d) Segmentation by the proposed method.}
\label{fig:segmentation}
\vskip -4mm
\end{figure}

A silhouette is a binary image that is obtained by separating the observed objects from the background.
Since player occlusion often occurs during sports events, the accuracy of the existing segmentation algorithms fails to meet the demands of FVV production.
To solve this problem, we propose an adaptive background subtraction method. 
Our method first performs Mask R-CNN \cite{he2017mask} to predict objects{'} silhouettes in an image and then generates a distance map that represents the shortest distance from a pixel to the predicted region.
In the next step, we extract {objects'} silhouettes using a background subtraction method \cite{backgroundModeling} in which thresholds for separation are adaptively updated according to the shortest distance in the distance map.
Fig.~\ref{fig:segmentation} (a) shows the foreground region extracted by Mask R-CNN in which some objects are not segmented correctly.
Fig.~\ref{fig:segmentation} (b) is a distance map where the darker color represents the distance {nearer} to the predicted silhouettes.
Fig.~\ref{fig:segmentation} (c) shows the segmentation results obtained by the conventional method \cite{backgroundModeling}. 
It can be seen that the spectators in the stand are separated from {the} background because their poses change dynamically during recording.
Fig.~\ref{fig:segmentation} (d) presents the segmentation results {obtained using} our proposed method.
It can be seen that the noise in the conventional method is removed while the missing parts in Mask R-CNN are recovered.

\subsection{Volumetric Visual Hull Reconstruction}

To approximate the 3D shape of an observed object, the volumetric visual hull first discretizes a pre-defined 3D volume into voxels, and then test{s} whether a voxel is occupied or not by projecting it onto all the silhouettes.
The voxel {which} falls outside the silhouettes is considered {to be an} unoccupied one.  
While it is robust and efficient, the voxel density in a pre-defined 3D volume seriously affects the accuracy of a visual hull.
A higher density produces a better shape approximation.   
However, along with the increasing of voxel density, the memory consumption and execution time also increase sharply. 
To solve these problems, we propose a {coarse-to-fine} visual hull construction method {in which a rough 3D shape approximation is carried out with low-resolution voxels and then an accurate reconstruction is performed on the ROIs with high-resolution voxels}.

\subsubsection{Visual Hull Reconstruction with Sparse Voxels}
In this step, we approximate the 3D shape of an observed scene with sparse voxels defining the whole scene as a pre-defined 3D volume.
Fig.~\ref{fig:sparse_visual_hull} (a) demonstrates a sparse reconstruction for a volleyball {sequence} in which the interval of voxels along the $x,y,z-$direction is $50\,mm$, while the number of voxels for occupy testing and occupied are $2.3 \times 10^7$ and $9.6 \times 10^3$, respectively.

\subsubsection{Noise Filtering and 3D ROI Extraction}
\begin{figure}[t]
\footnotesize
\centering
\includegraphics[width=0.60\linewidth]{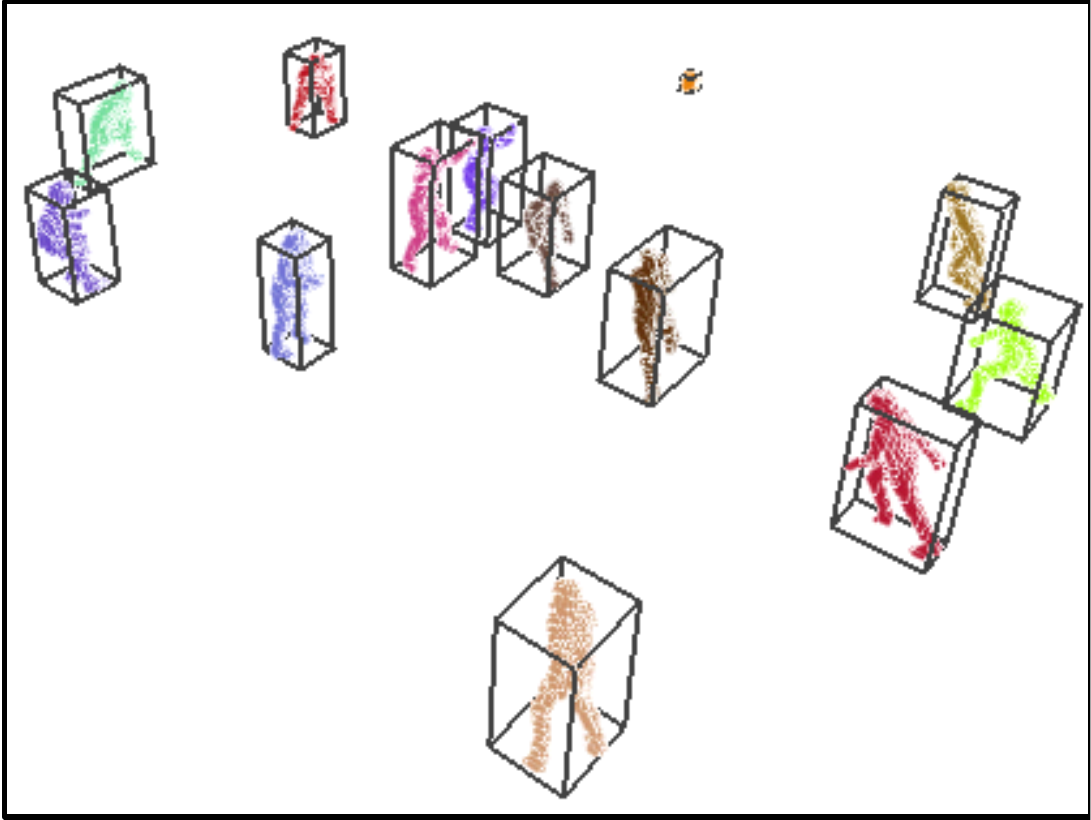}
\caption{3D ROI extraction for sparse point cloud.}
\label{fig:ccl}
\vskip -2mm
\end{figure}

Once the sparse volumetric visual hull is obtained, the individual objects are clustered using a connected components labeling algorithm \cite{8476292}.
The original algorithm is designed for a 2D image.
Here, we extend it to 3D space.
We express the pre-defined 3D volume as a binary volume in which only the occupied voxel is denoted as ON state.
The volume is divided into independent blocks, and each block is assigned to different GPU processors to perform local and global labeling.
It should {be noted} that $26$-adjacency is used in both {the} local and global label stages.
The minimum 3D bounding box for a connected point set is found by traversing the final label map. 
Fig.~\ref{fig:ccl} shows a clustering result in which each object is assigned  a unique color {as well as showing} the minimum bounding box of each object.
The reconstructed visual hull may {contain} noise that comes from imperfect silhouettes.
To remove noise, we {establish} a criterion {taking into account} the number of voxels in one point set as expressed in {{Eq.}~(\ref{equ:noisefilter})}.
\begin{eqnarray}
\mathcal{S}_{t} =
\begin{cases} 
OFF,  & \mbox{if }  T_{nb}<N(\mathcal{S}_{t})<T_{np} \\
ON, & \mbox{otherwise}
\end{cases} {.}
\label{equ:noisefilter}
\end{eqnarray}
Here, $N(\mathcal{S}_{t})$ expresses the number of voxels in the $t-$th point set $\mathcal{S}_{t}$.
We remove a point set if the number of voxels of the point set is less than a specified voxel number $T_{np}$ or larger than another specified voxel number $T_{nb}$.

\subsubsection{Visual Hull Reconstruction with Dense Voxels}
\begin{figure}[t]
\centering
\footnotesize
	\begin{minipage}[b]{0.84\linewidth}
 		\centering
 		\subfloat[Sparse volumetric visual hull]
		{
 	 		\begin{overpic}[width=1\textwidth]
 	 			{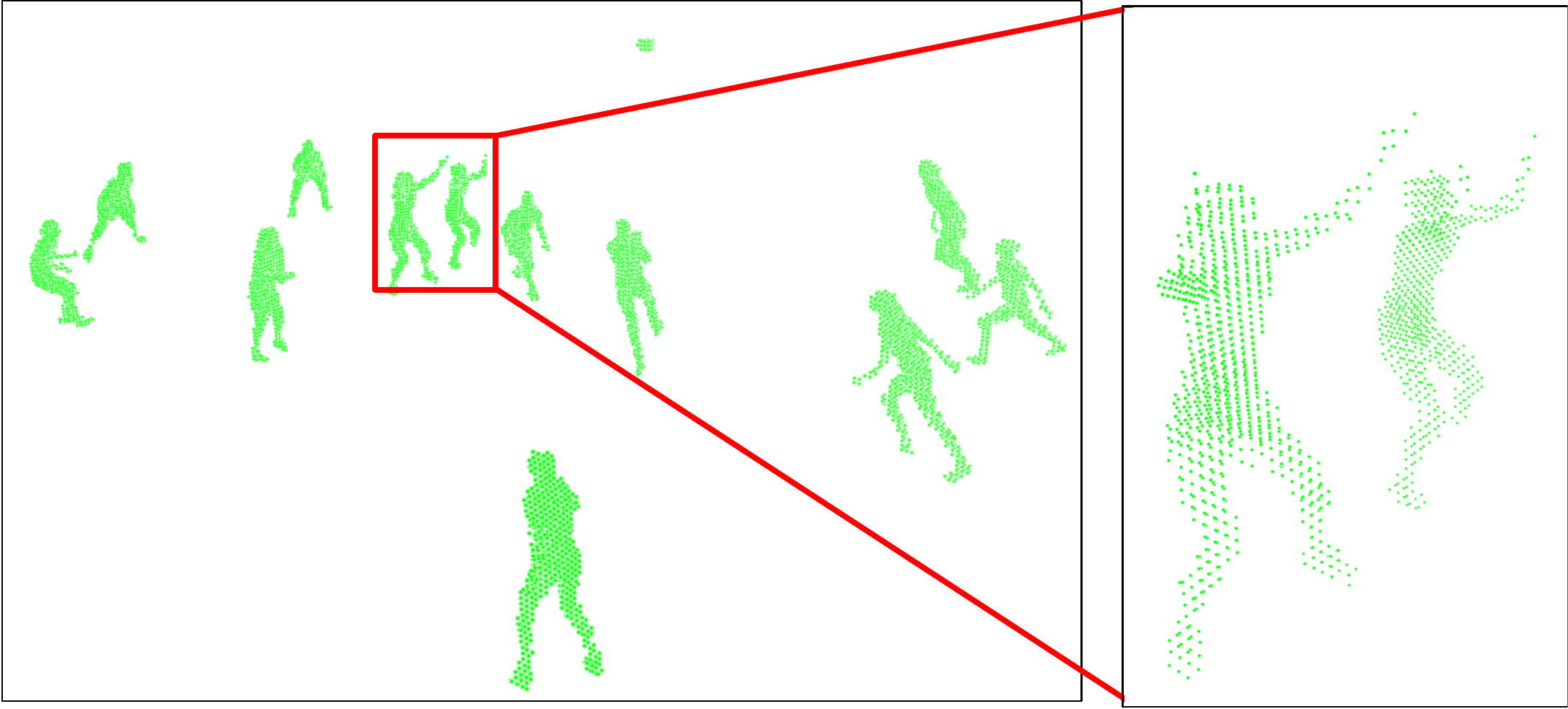}
 		\end{overpic}
 	 	}
	\end{minipage}
\vskip 1mm
	\begin{minipage}[b]{0.84\linewidth}
 		\centering
 		\subfloat[{Dense} volumetric visual hull]
		{
 			\begin{overpic}[width=1\textwidth]
 	 			{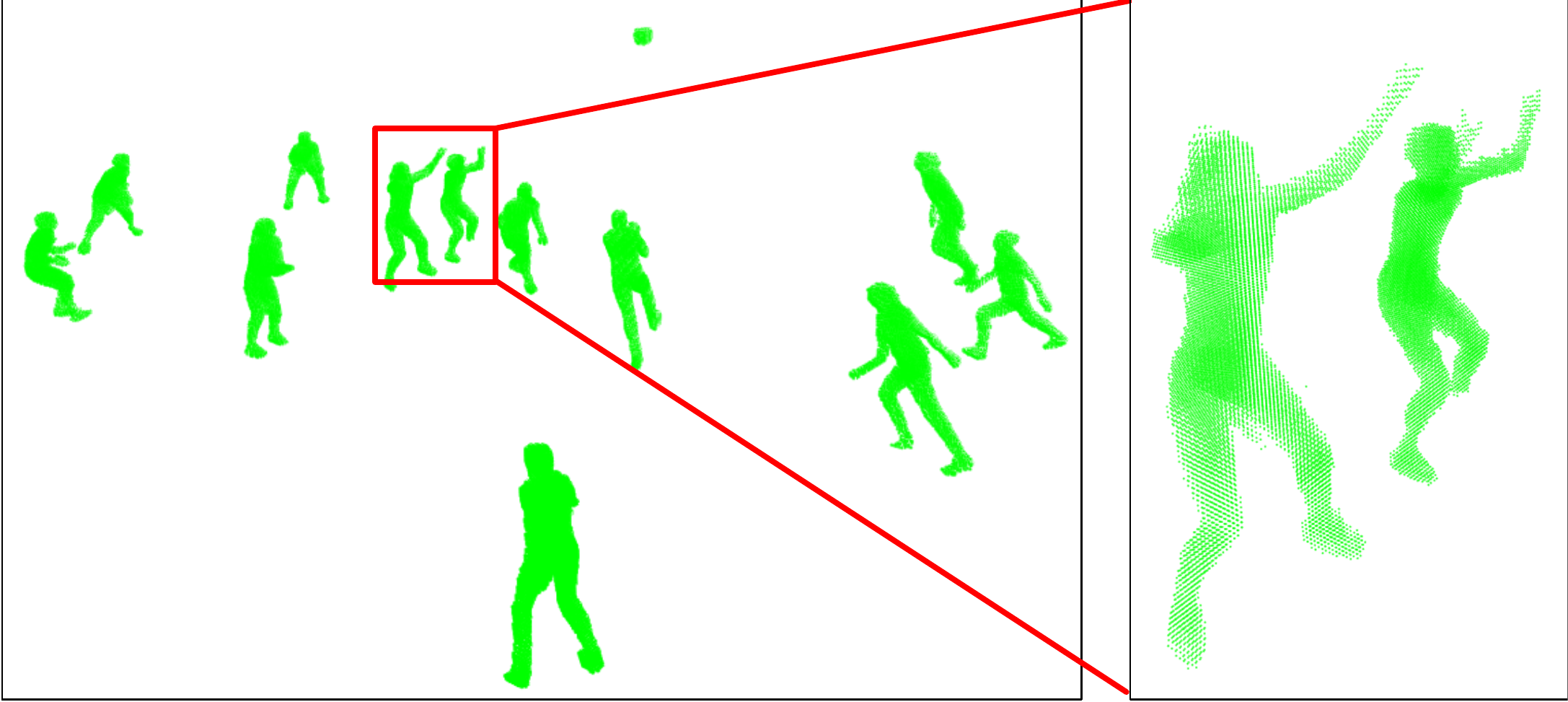}
 		\end{overpic}
 		}
	\end{minipage}	
\caption{Volumetric visual hull.}
\label{fig:sparse_visual_hull}
\vskip -4mm
\end{figure}

In this step, we construct a high-density point cloud for each object considering each 3D ROI as an individual pre-defined 3D volume.
Fig.~\ref{fig:sparse_visual_hull} (b) shows the dense reconstruction for {the same volleyball sequence} in which the interval of voxels along the $x,y,z-$direction is $20\,mm$, while the number of voxels for occupy testing and occupied is $3.6 \times 10^7$ and $1.4 \times 10^5$, respectively.
{By comparing Fig.~\ref{fig:sparse_visual_hull} (a) and (b), it can be seen that the proposed method increases the number of occupied voxels around $15$-fold in {the} case {where} the number of voxels for testing is similar.}

\subsection{Surface polygonization}

\begin{figure}[t]
\centering
\footnotesize
	\begin{minipage}[b]{0.6\linewidth}
 		\centering
 		\subfloat[]
		{
 	 		\begin{overpic}[width=1\textwidth]
 	 			{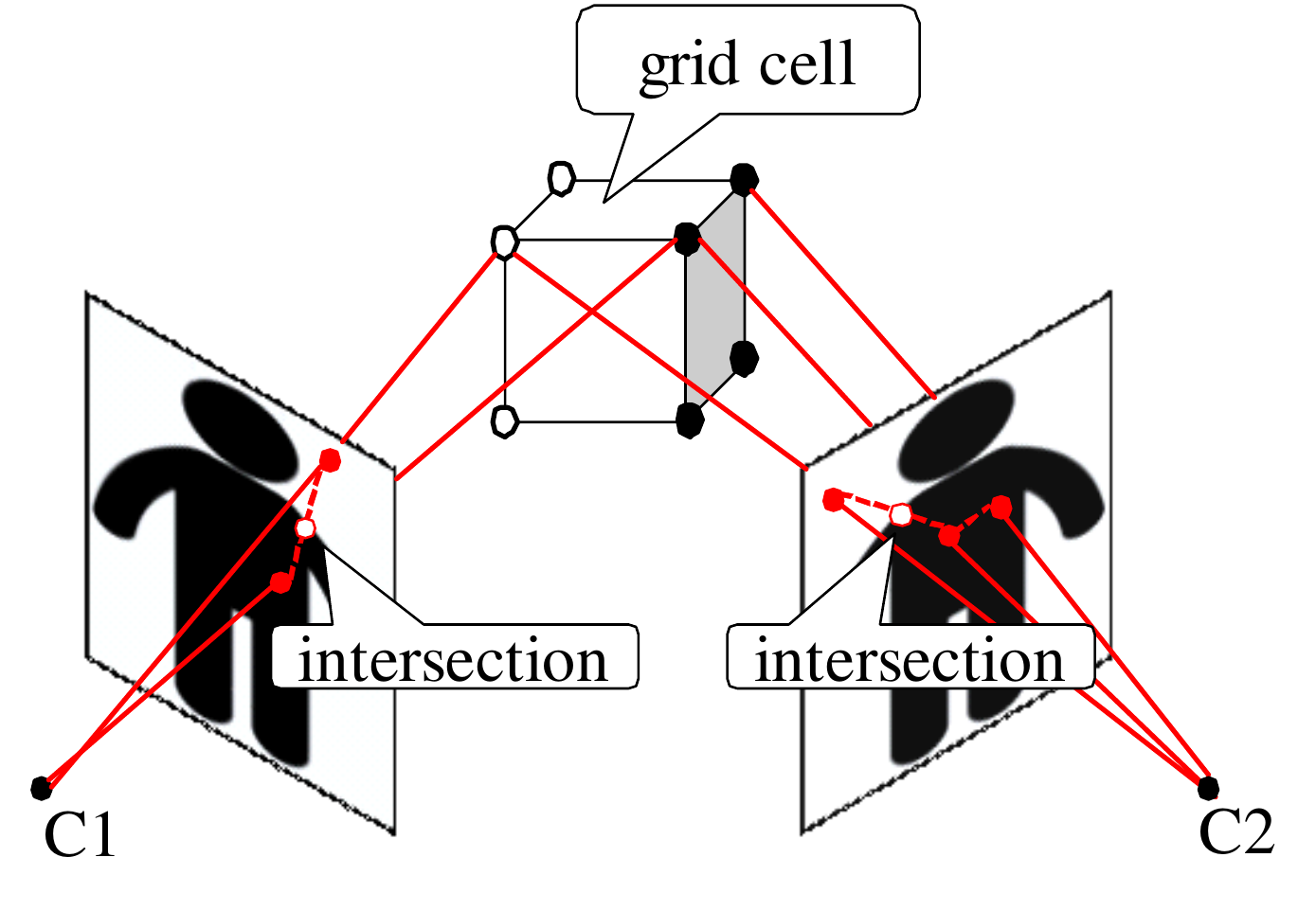}
 		\end{overpic}
 	 	}
	\end{minipage}
	\begin{minipage}[b]{0.38\linewidth}
 		\centering
 		\subfloat[]
		{
 			\begin{overpic}[width=1\textwidth]
 	 			{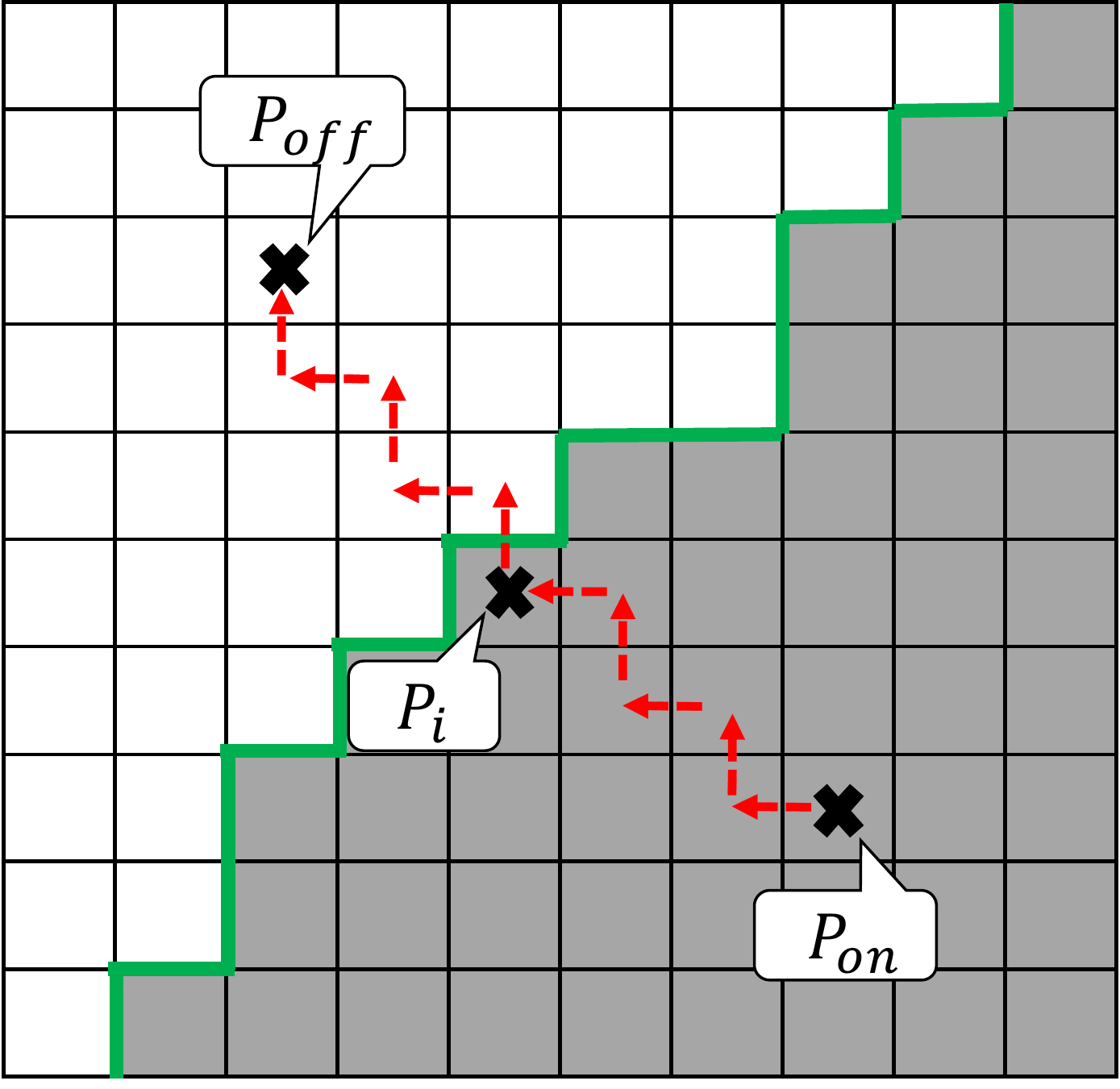}
 		\end{overpic}
 		}
	\end{minipage}	
\caption{Exact isovalue computation. (a) {Project the edges of a grid cell onto image plane.} (b) Find the exact intersection using Bresenham's line algorithm.}
\label{fig:isovalue}
\vskip -4mm
\end{figure}

The volumetric visual hull can be represented by a set of grid cells, in which the eight vertices of a cell may be in same or different states.
Cells with different vertex states intersect with the visual hull, while the others are inside or outside the visual hull.
The intersection, also called isosurface, cuts the edge{, the two endpoints of which have different states.}
To obtain the exact isosurface, we project the intersected edge onto each image plane as demonstrated in Fig.~\ref{fig:isovalue} (a).
The projection of the vertex with ON state $P_{on}$ falls in the foreground in all the silhouettes, while the projection of the other vertex $P_{off}$ falls in the background in at least one silhouette.
For a specific silhouette, We find the exact intersection between {the} projection line with {a} silhouette using Bresenham's line algorithm.
The last foreground pixel when traversing the projection line is considered {to be} intersection pixel denoted as $P_i$ (shown in Fig.~\ref{fig:isovalue} (b)).
{T}he isovalue $\lambda_i$ in camera $i$ can be calculated by the following equation:
\begin{eqnarray}
\lambda_i = \frac{\left \| P_i-P_{on}\right \| }{\left \| P_{off}-P_{on}\right \| } {.}
\end{eqnarray}
To guarantee the polygonal mesh is the maximum approximation of object{'s} shape, we define the isovalue $\lambda$ for a grid cell edge as the minimum of intersections $\lambda_i (i=1{,}\cdots,N)$ in all cameras as {represented by} the following equation. 
{$N$} is the number of cameras.
\begin{eqnarray}
\lambda = \min \left\{ \lambda_1, \lambda_2, \cdots, \lambda_N \right\} {.}
\label{equ:isovalueDetermination}
\end{eqnarray}
Once the isovalue for each grid cell is obtained, the isosurface is built by following the configurations of the marching cubes algorithm.
Finally, we express the surface using triangles.

\subsection{Visibility Detection}

To determine the visibility of a specific camera $i$, we perform two operations including the computation of the depth image and determination of occlusion.
\subsubsection{Computation of depth image}
We project each triangle on an object surface onto the image plane of camera $i$ to form a 2D triangle.
The depth ${}^iD^{j}_m$ of a pixel $j$ that is bounded by the $m-$th 2D triangle is assigned the distance from its corresponding 3D triangle to the camera center.
After projecting all the triangles, the depth ${}^iD^{j}$ of a pixel $j$ is determined by {Eq.~(\ref{equ:isovalueDetermination})}.
\begin{eqnarray}
{}^iD^{j} = \min \left\{ {}^iD^{j}_1, {}^iD^{j}_2, \cdots, {}^iD^{j}_M \right\} {.}
\label{equ:isovalueDetermination}
\end{eqnarray}
Here, {$M$} expresses the number of 3D triangles corresponding {{with}} pixel $j$.
The upper image in Fig.~\ref{fig:flowchart} (d) shows the computed depth image, in which the darker color {{indicates that}} the distance to the camera center {{is greater}}.
{{T}}o find the pixels bounded by a triangle, we assume that a triangle is orientable and the bounded pixel is on the same side of each edge.

\subsubsection{Visibility detection}
The visibility of each triangle in each camera is tested by comparing the distance from the triangle to a camera plane with the cached value in {{the}} depth image.
If the difference is larger than a threshold $T_v$, we consider that the triangle is occluded.
The bottom image in Fig.~\ref{fig:flowchart} (d) presents an occlusion map of a specific camera where the occluded parts are {{shown in}} grey.
It should be noted that $T_v$ is assigned a high value in our implementation so that the self-occlusion is ignored.

\subsection{View-dependent Rendering}

When users experience FVV, the 3D coordinate and the direction of a virtual viewpoint are calculated.
The uppermost reference camera for rendering is identified as the nearest camera by calculating the distance from the virtual viewpoint to each camera.
Coupled with the occlusion maps, the non-occluded parts are rendered by the uppermost reference camera, while {the} neighboring cameras render the occluded regions.


\begin{figure}[t]
\centering
\footnotesize
	\begin{minipage}[b]{0.49\linewidth}
 		\centering
 		\subfloat[{Camera configuration}]
		{
 	 		\begin{overpic}[width=1\textwidth]
 	 			{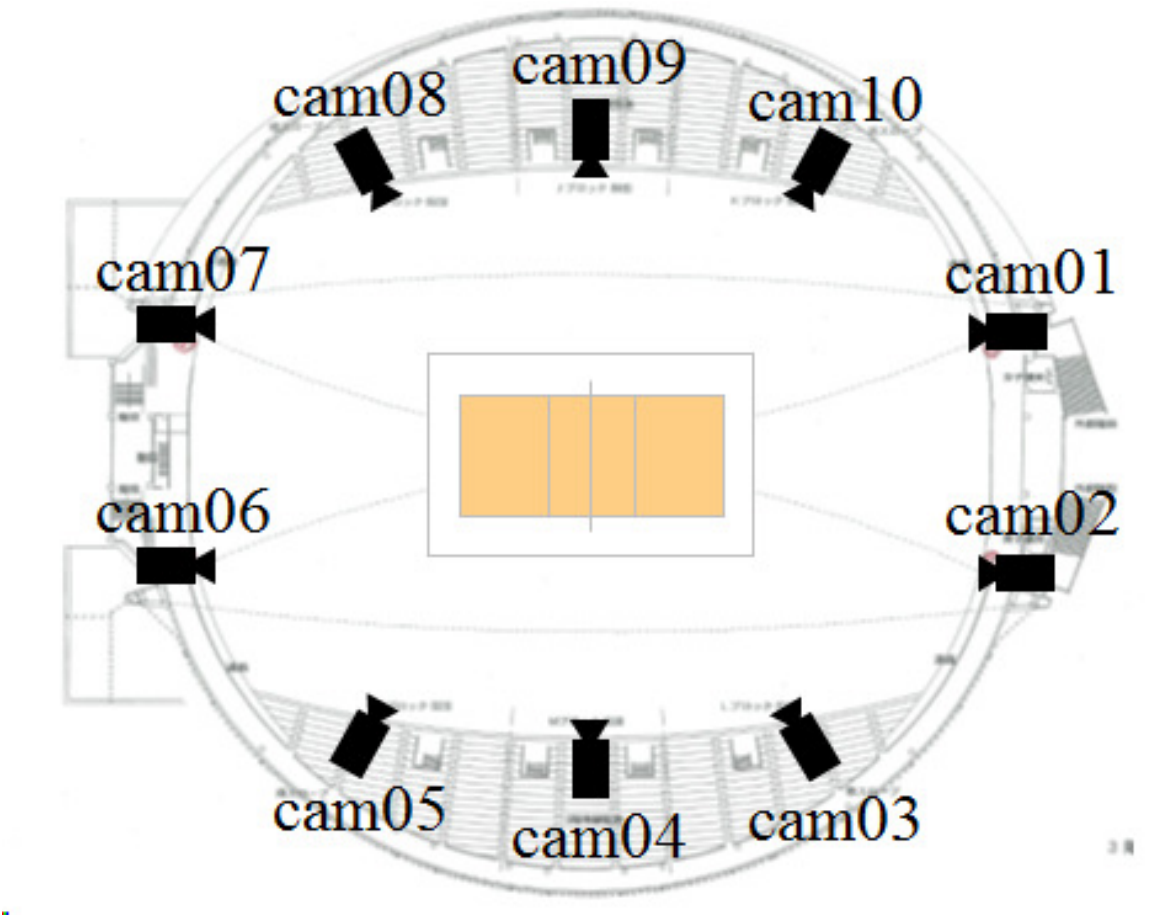}
 		\end{overpic}
 	 	}
	\end{minipage}
	\begin{minipage}[b]{0.49\linewidth}
 		\centering
 		\subfloat[{Recording environment}]
		{
 			\begin{overpic}[width=1\textwidth]
 	 			{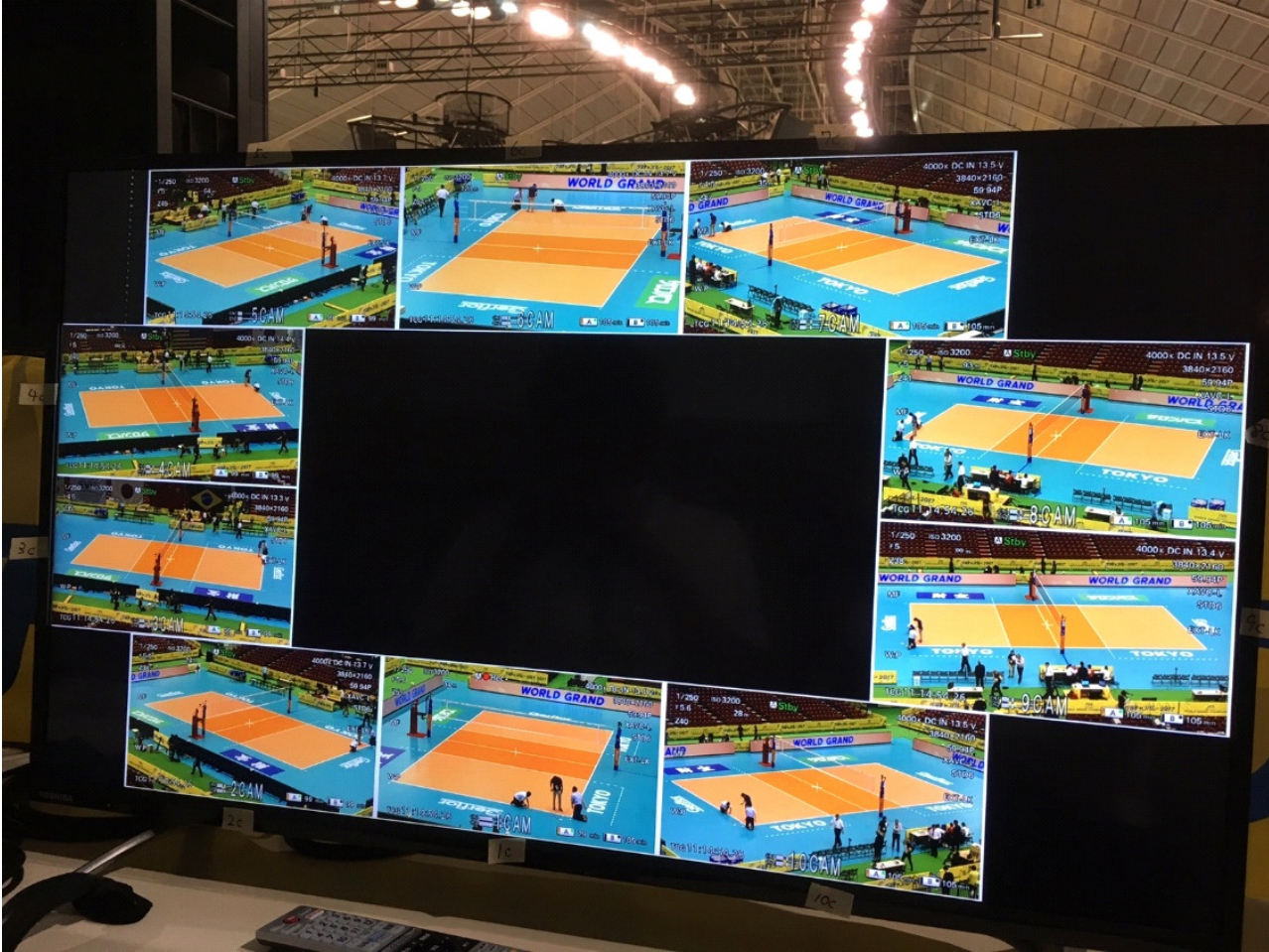}
 		\end{overpic}
 		}
	\end{minipage}	
\caption{{System overview.}}
\label{fig:camera-position}
\vskip -0mm
\end{figure}

\begin{figure}[t]
\centering
\footnotesize
	\begin{minipage}[b]{0.49\linewidth}
 		\centering
 		\subfloat[The first content]
		{
 	 		\begin{overpic}[width=1\textwidth]
 	 			{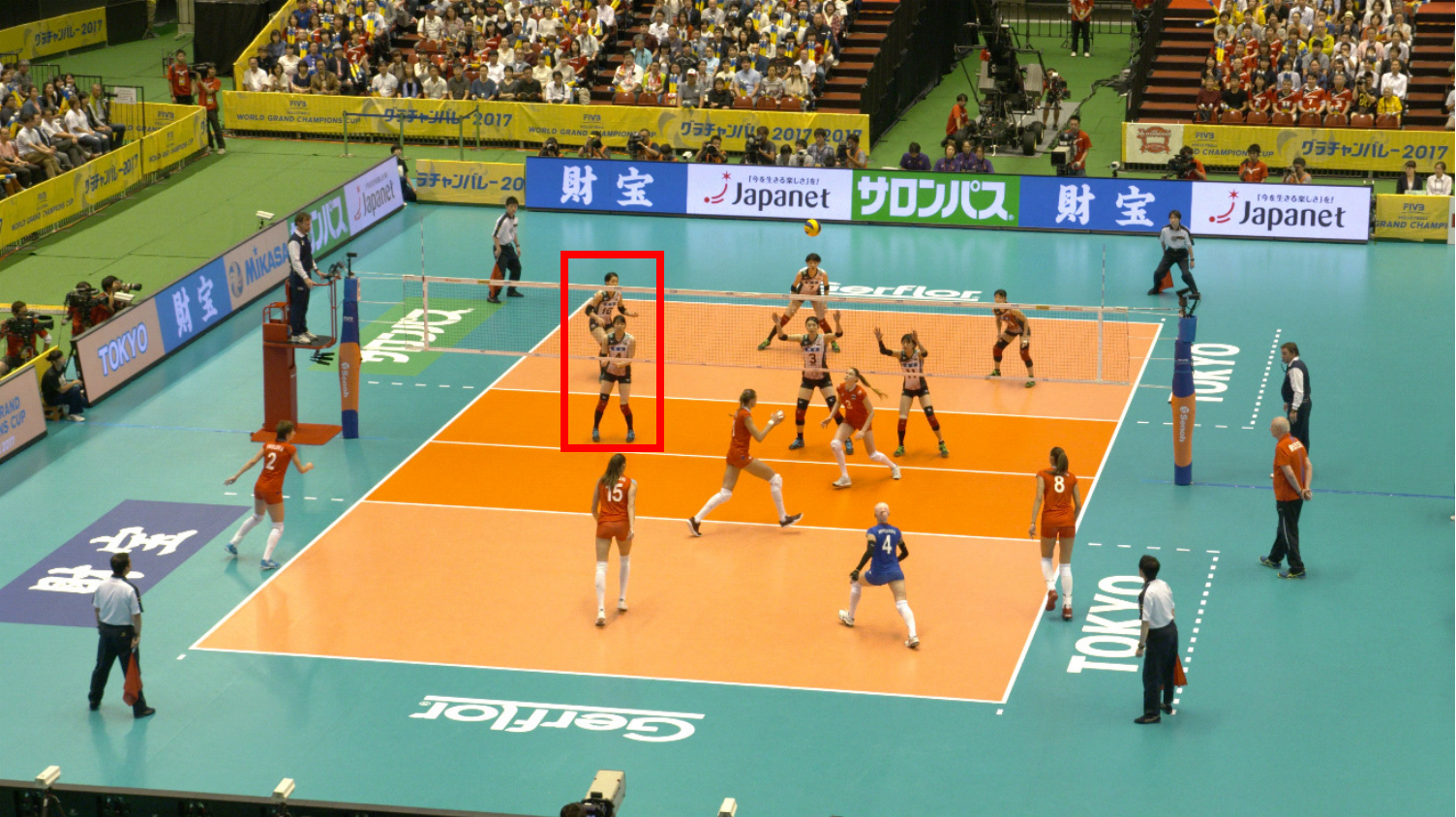}
 		\end{overpic}
 	 	}
	\end{minipage}
	\begin{minipage}[b]{0.49\linewidth}
 		\centering
 		\subfloat[The second content]
		{
 			\begin{overpic}[width=1\textwidth]
 	 			{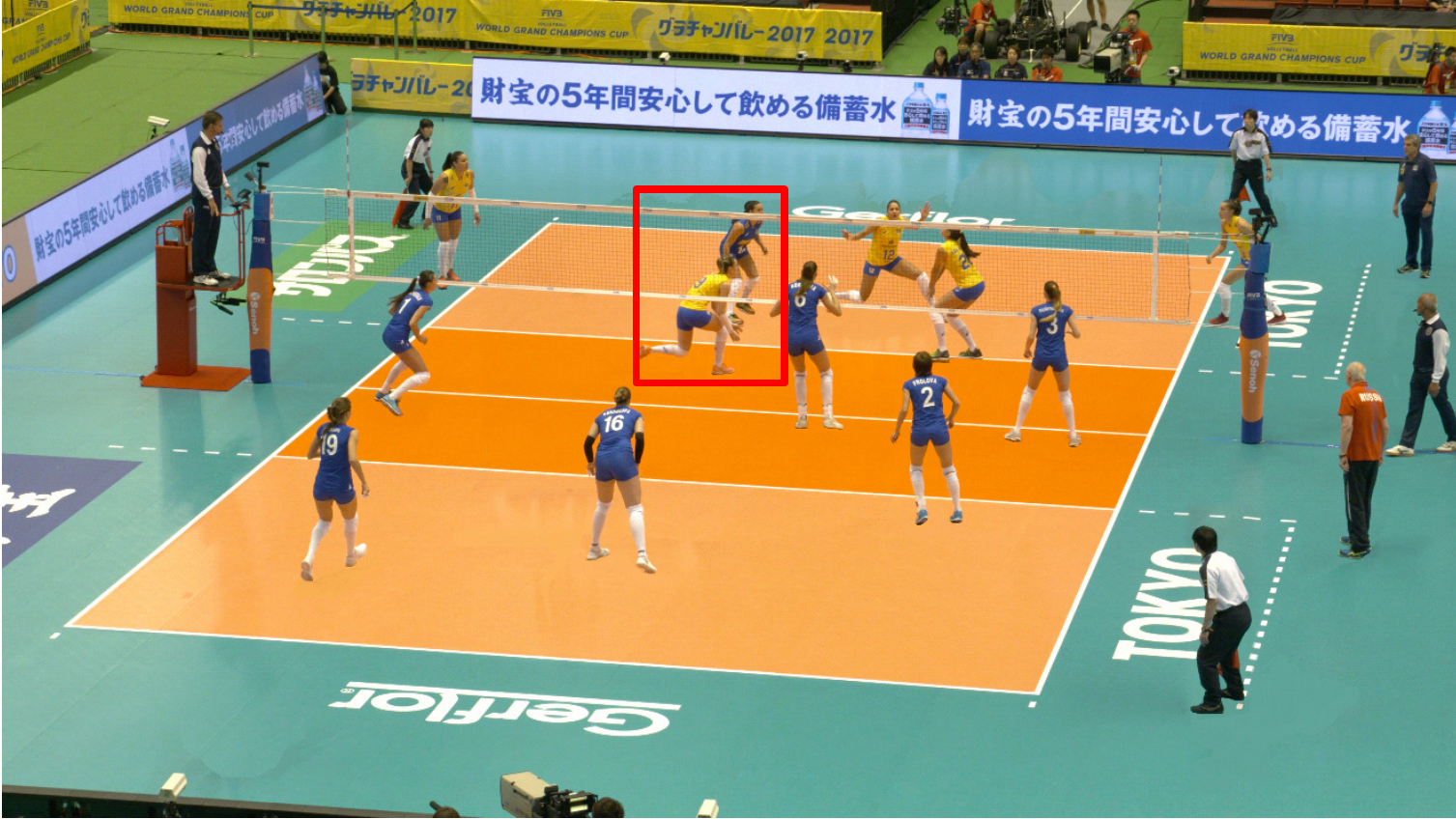}
 		\end{overpic}
 		}
	\end{minipage}	
\caption{{The input image of the Volleyball contents.}}
\label{fig:the-input-image}
\vskip -4mm
\end{figure}

\section{EXPERIMENTS}

\begin{figure*}[t]
\centering
\footnotesize
	\begin{minipage}[b]{0.32\linewidth}
 		\centering
 		\subfloat[]
		{
 	 		\begin{overpic}[width=1\textwidth]
 	 			{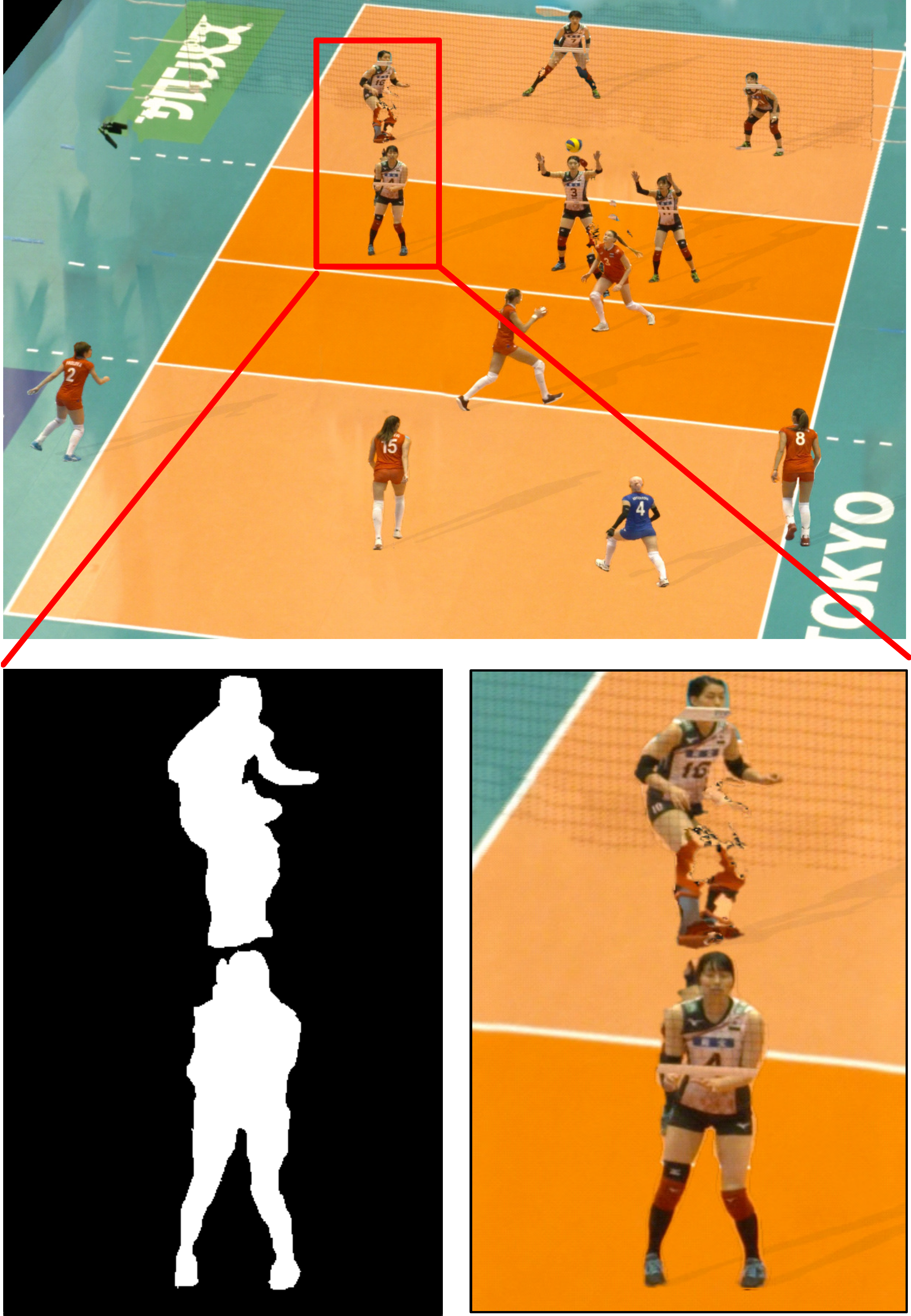}
 		\end{overpic}
 	 	}
	\end{minipage}
\hskip 1mm
	\begin{minipage}[b]{0.32\linewidth}
 		\centering
 		\subfloat[]
		{
 			\begin{overpic}[width=1\textwidth]
 	 			{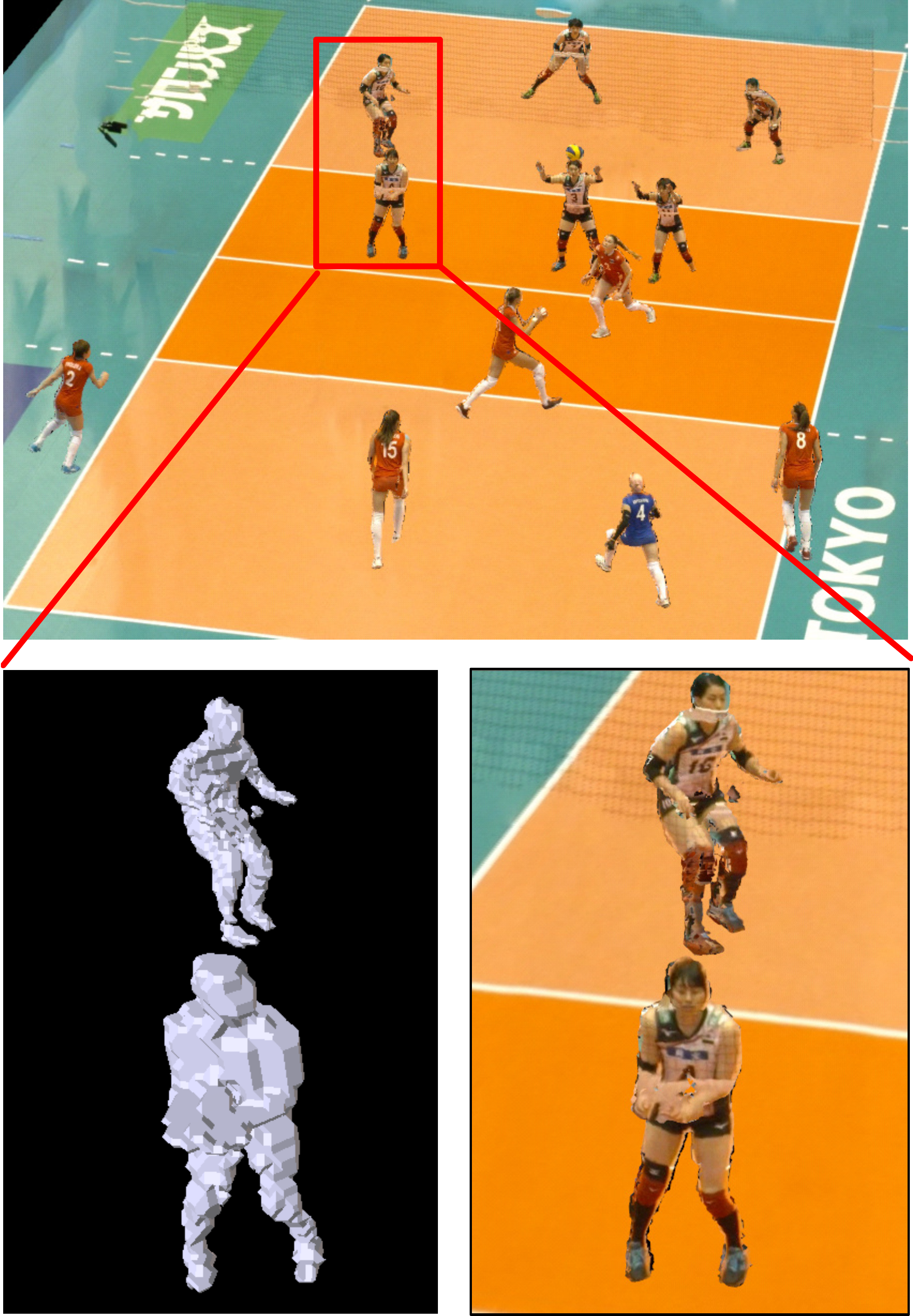}
 		\end{overpic}
 		}
	\end{minipage}	
\hskip 1mm
	\begin{minipage}[b]{0.32\linewidth}
 		\centering
 		\subfloat[]
		{
 			\begin{overpic}[width=1\textwidth]
 	 			{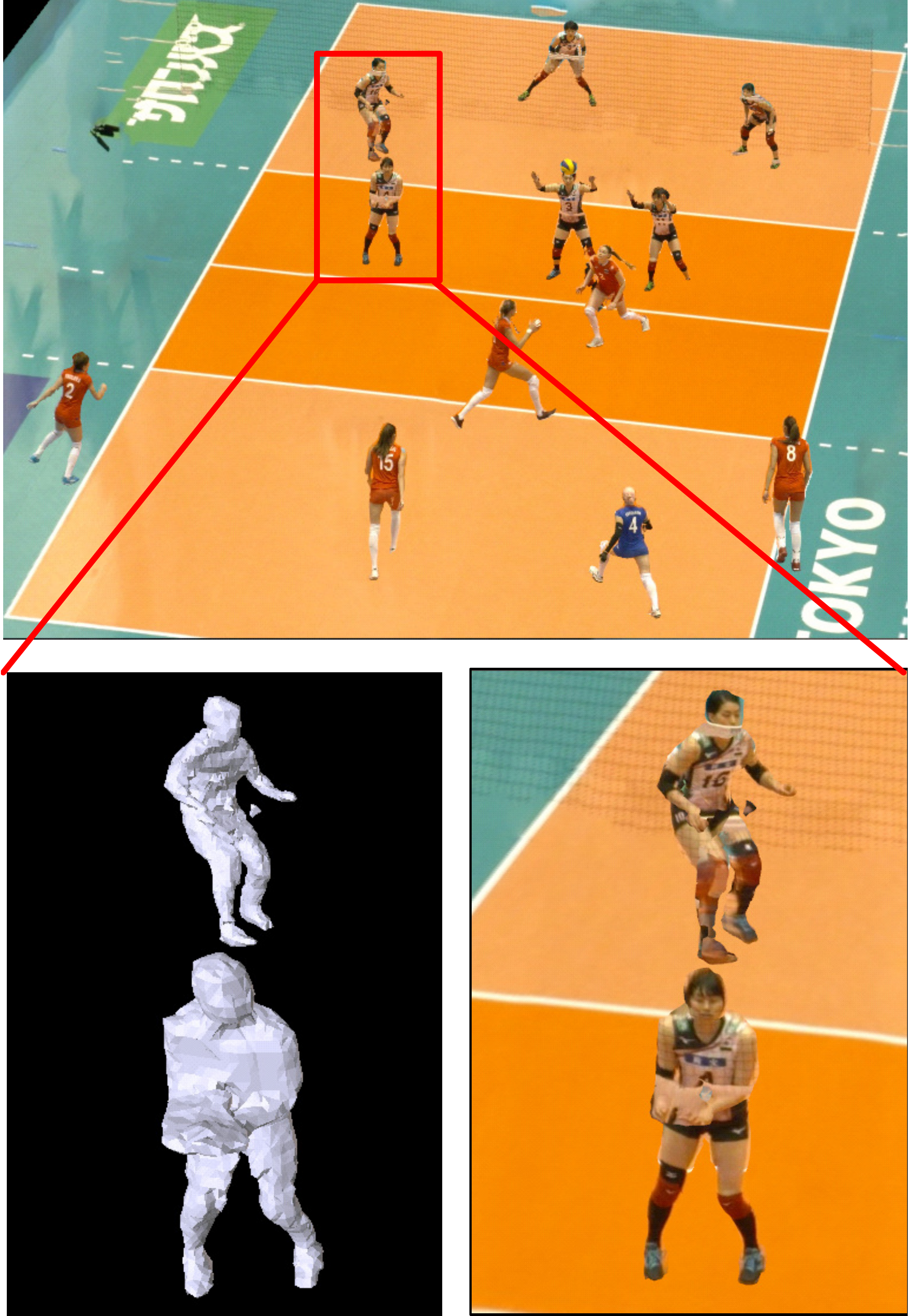}
 		\end{overpic}
 		}
	\end{minipage}		
\caption{Synthesized FVV of the first volleyball sequence viewing from a virtual viewpoint. (a) result by \cite{Sankoh2018Acmmm}. (b) result by \cite{smolic20113d}. (c) result by our method. The left bottom figure of (a) is reconstructed binary 2D billboard while the left bottom figure of (b) and (c) are reconstructed 3D polygonal mesh.}
\label{fig:systhesis-results-1}
\vskip -4mm
\end{figure*}

\begin{figure*}[t]
\centering
\footnotesize
	\begin{minipage}[b]{0.32\linewidth}
 		\centering
 		\subfloat[]
		{
 	 		\begin{overpic}[width=1\textwidth]
 	 			{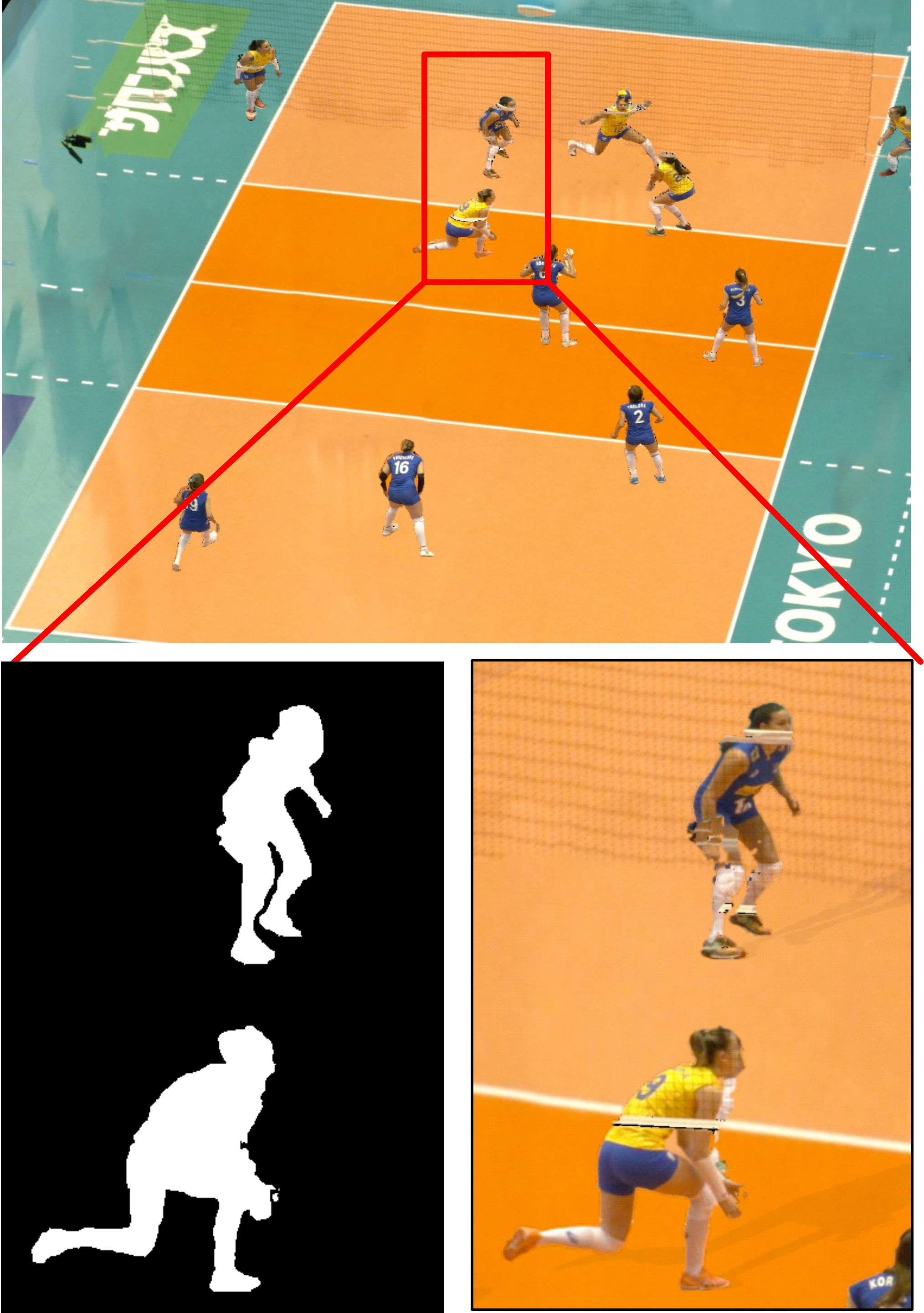}
 		\end{overpic}
 	 	}
	\end{minipage}
\hskip 1mm
	\begin{minipage}[b]{0.32\linewidth}
 		\centering
 		\subfloat[]
		{
 			\begin{overpic}[width=1\textwidth]
 	 			{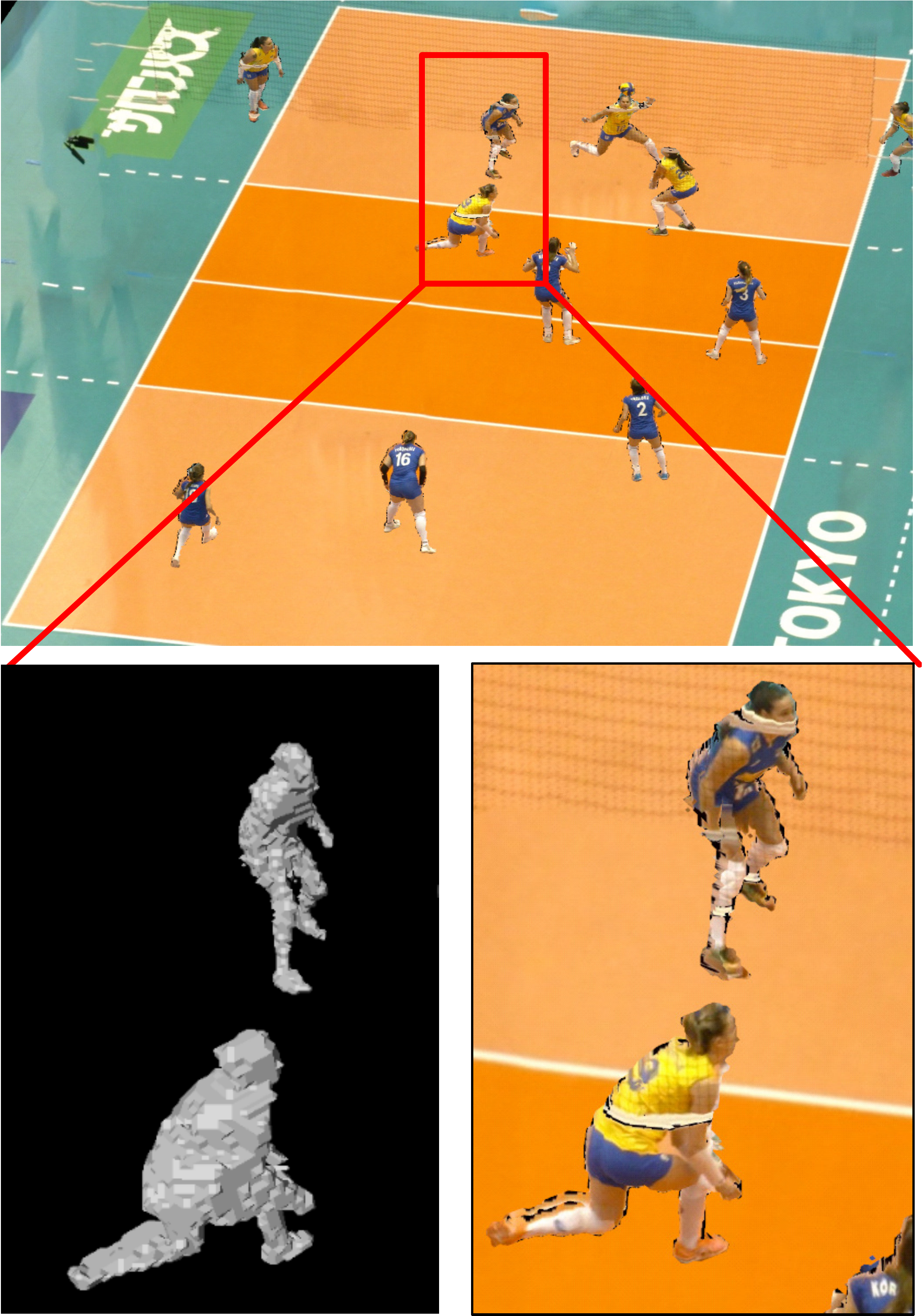}
 		\end{overpic}
 		}
	\end{minipage}	
\hskip 1mm
	\begin{minipage}[b]{0.32\linewidth}
 		\centering
 		\subfloat[]
		{
 			\begin{overpic}[width=1\textwidth]
 	 			{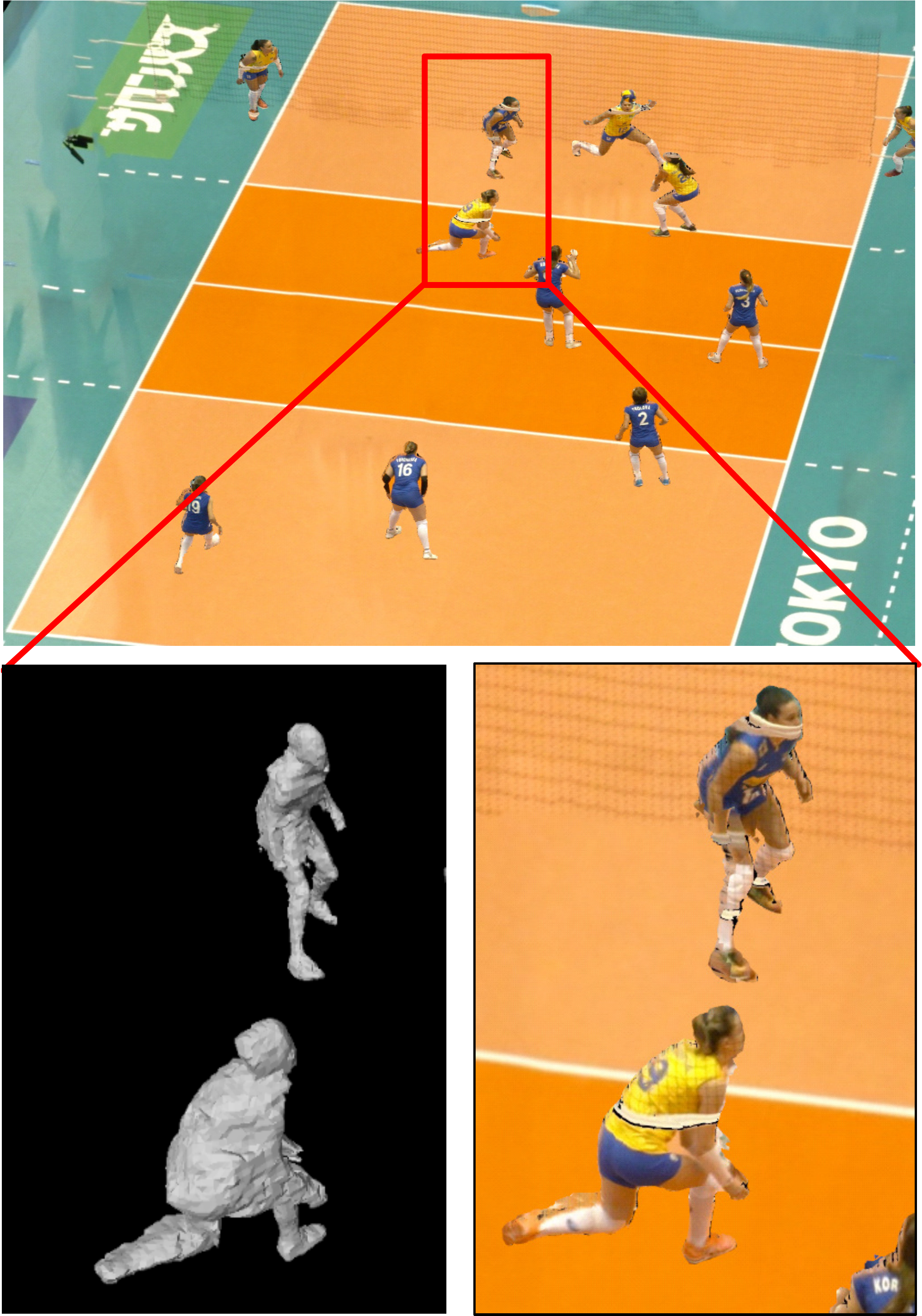}
 		\end{overpic}
 		}
	\end{minipage}		
\caption{{Synthesized FVV of the second volleyball sequence viewing from a virtual viewpoint. (a) result by \cite{Sankoh2018Acmmm}. (b) result by \cite{smolic20113d}. (c) result by our method. The left bottom figure of (a) is reconstructed binary 2D billboard while the left bottom figure of (b) and (c) are reconstructed 3D polygonal mesh.}}
\label{fig:systhesis-results-2}
\vskip -4mm
\end{figure*}

\begin{figure*}[t]
\centering
\footnotesize
	\begin{minipage}[b]{0.97\linewidth}
 		\centering
 		\subfloat[Images captured at different time by a selected camera]
		{
 	 		\begin{overpic}[width=1\textwidth]
 	 			{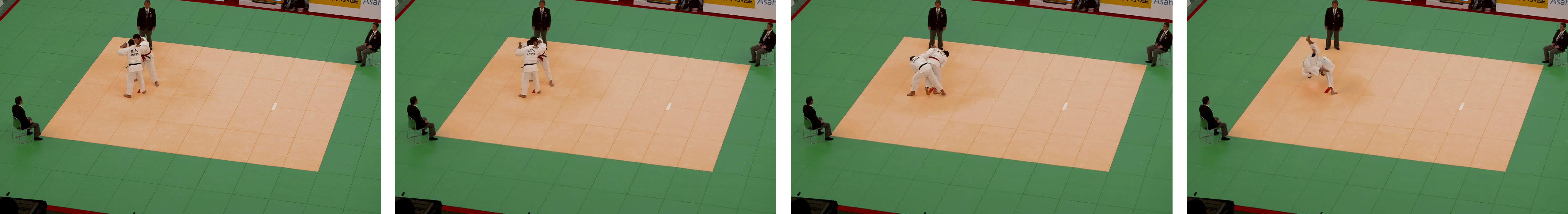}
 		\end{overpic}
 	 	}
	\end{minipage}
\vskip 2mm
	\begin{minipage}[b]{0.97\linewidth}
 		\centering
 		\subfloat[Synthesised images by \cite{Sankoh2018Acmmm}]
		{
 			\begin{overpic}[width=1\textwidth]
 	 			{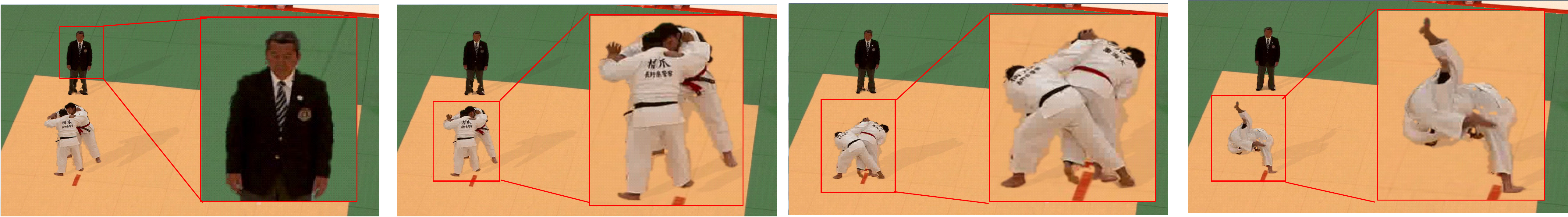}
 		\end{overpic}
 		}
	\end{minipage}	
\vskip 2mm
	\begin{minipage}[c]{0.97\linewidth}
 		\centering
 		\subfloat[Synthesised images by \cite{smolic20113d}]
		{
 			\begin{overpic}[width=1\textwidth]
 	 			{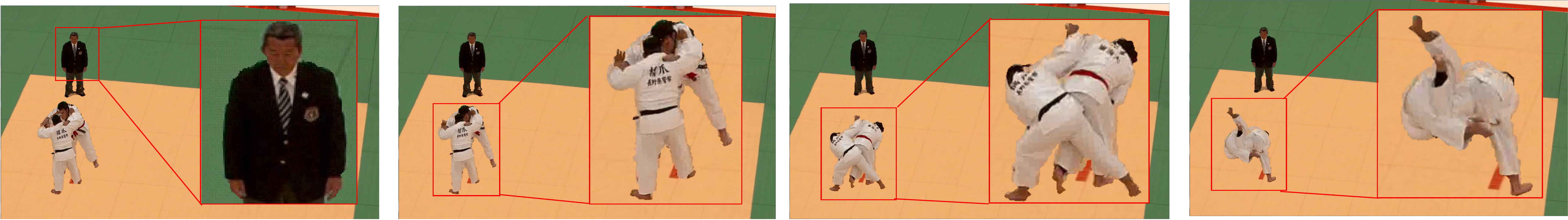}
 		\end{overpic}
 		}
	\end{minipage}	
\vskip 2mm
	\begin{minipage}[d]{0.97\linewidth}
 		\centering
 		\subfloat[Synthesised images by the proposed method]
		{
 			\begin{overpic}[width=1\textwidth]
 	 			{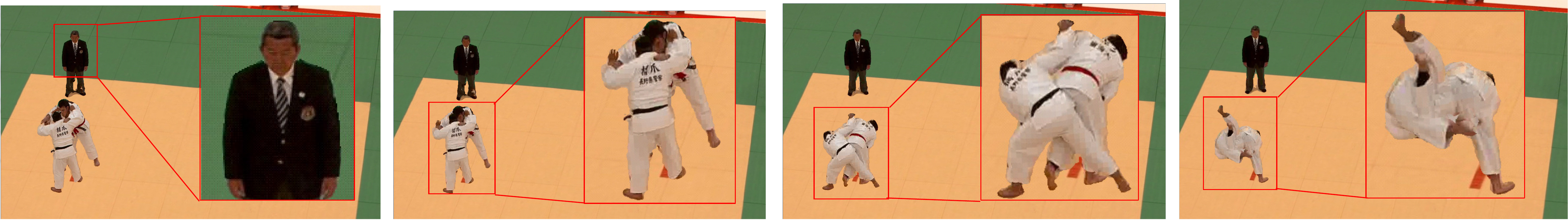}
 		\end{overpic}
 		}
	\end{minipage}			
\caption{Synthesized FVV of a judo sequence with a virtual stadium model}
\label{fig:systhesis-results-Judo}
\vskip -4mm
\end{figure*}

\subsection{Visual quality evaluation}

To demonstrate the performance of our method, we compared it to two approaches. 
The first one \cite{Sankoh2018Acmmm} is a billboard-based method that reconstructs a billboard for each object including an occluded {object} by employing a rough 3D model.
The second one \cite{smolic20113d} is a typical 3D model-based method that approximates the object shape using visual hull and {builds} the mesh representation using the conventional marching cubes algorithm.

We {initially} evaluated the proposed method with volleyball {sequences} that are captured with ten synchronized cameras.
The resolution of each camera {wa}s $3840\times2160$, and the frame rate is $30~$fps.
The target space for reconstruction was set to $18~$meters wide, $18~$meters deep, and $9~$meters high.
Fig.~\ref{fig:camera-position} shows the camera configuration in the stadium and { recording environment}.
The camera threshold in \cite{Sankoh2018Acmmm} for the reconstruction of a rough 3D shape is $9$, while the isovalue for isosurface extraction in \cite{smolic20113d} is $0.5$.
The voxel size in all the three methods is $2\,cm$ $\times$ $2\,cm$ $\times$ $2\,cm$. 
Fig.~\ref{fig:the-input-image} shows the input images of two volleyball {contents} captured by camera $01$.
It can be seen that the players in the red rectangular are occluded.
Fig.~\ref{fig:systhesis-results-1} and \ref{fig:systhesis-results-2} demonstrate virtual viewpoint images generated by the three methods. 
The top row shows the synthesized representations of the whole scene, 
{while the bottom row presents reconstructed models (2D billboard or 3D polygon mesh) and close-up views.}
The virtual viewpoint images are rendered mainly by camera $01$ because it is the camera nearest to the virtual viewpoint.
{First, let us focus on the results by \cite{Sankoh2018Acmmm} (shown in Fig.~\ref{fig:systhesis-results-1} (a) and \ref{fig:systhesis-results-2} (a)). 
The binary 2D billboard on the left bottom illustrates that this method can estimate the pose of an object reasonably, while the close-up view on the right bottom shows that some occluded regions failed to be rendered.
Next, let us look at the results by \cite{smolic20113d} (shown in Fig.~\ref{fig:systhesis-results-1} (b) and \ref{fig:systhesis-results-2} (b)).
We found that \cite{smolic20113d}  produces jagged artifacts in the surface of the reconstructed polygonal mesh, which affect the quality of rendering.
From the close-up view, it can be seen that black noise exists in the object boundary.
Furthermore, the appearance of the occluded area is misaligned.
As for the results by our approach (shown in Fig.~\ref{fig:systhesis-results-1} (c) and \ref{fig:systhesis-results-2} (c)), it can be observed that the reconstructed polygonal mesh is smooth, while the color appearance of occluded regions was appropriately reproduced.
}

A judo {sequence} that is captured with sixteen synchronized cameras was also employed to evaluate the visual quality of virtual images {synthesized using} the proposed method.
The resolution of captured images is $1920 \times 1080$ while the predefined 3D volume is $5\,m$ $\times$ $6\,m$ $\times$ $2.1\,m$.
{For all the methods, voxel size for 3D reconstruction is kept at $4\,cm$ $\times$ $4\,cm$ $\times$ $4\,cm$.}
The camera threshold {in \cite{Sankoh2018Acmmm}} and the isovalue {in \cite{smolic20113d}} are $14$ and $0.5$, respectively.
Fig.~\ref{fig:systhesis-results-Judo} (a) presents the input images of a selected camera, which were captured at different times.
Fig.~\ref{fig:systhesis-results-Judo} (b), (c), and (d) show the synthesized images {obtained} by the three methods with a virtual stadium model. 
{The red rectangular on the virtual playground is a point of reference, allowing people to sense the difference among the synthesized images.}
Concerning the results in (b), it can be seen that the position and shape of reconstructed objects are different from those obtained by 3D modeling methods even {though} their virtual viewpoints are the same.
The reason for this phenomenon is that the objects in (b) are represented by 2D billboards that cannot describe the exact 3D pose of each object.
{Regarding} the results in (c), the boundary of {the} reconstructed objects is jagged.
The imprecise polygonization in {\cite{smolic20113d}} decreases the realistic visual effect. 
{The observation from (c) is that the reconstructed model retains the same pose with those obtained by \cite{smolic20113d} while the object boundary is smooth and natural.}

\subsection{Execution time evaluation}

In the production of FVV, steps (B)--(D) are accelerated by executing them in parallel on a GPU board on the server side, while step (E), rendering, runs on the client side.
To achieve high computation{al} efficiency, we grab all the memory the production needs from GPU {immediately} at startup of the application.
The data transfer between CPU and GPU and processes for production are managed separately by two CPU threads.
The reference methods, \cite{Sankoh2018Acmmm} and \cite{smolic20113d}, are implemented in the same manner. 
We use a PC with Intel(R) Core(TM) i7-6700K CPU, 4.00 GHz $\And$ 4.00 GHz, 32.0 GB RAM, NVIDIA Geforce GTX 1080Ti and Windows 7 Professional Service Pack 1.

\begin{table}[t]
\caption{Execution time of production in milliseconds.}
\label{tab:time-1}
\centering
{\footnotesize
\begin{tabular}{|l|c|c|c|}
\hline
& \cite{Sankoh2018Acmmm} & \cite{smolic20113d} & proposed     \\
\hline\hline
1st Volleyball content & 15650.15 & 1411.59 & 36.39 \\
\hline
2nd Volleyball content & 14006.47 & 1436.17 & 33.66 \\
\hline
Judo content       & 956.58  & 17.73 & 7.30 \\
\hline
\end{tabular}
}
\vskip -4mm
\end{table}

Table~\ref{tab:time-1} shows the execution time for the processes in CUDA excluding the time {taken for} device memory allocation and data transfer when the voxel sizes for sparse and dense volumetric visual hull approximation for the three {sets of} content are respectively {$\left(50\,mm, 20\,mm\right)$, $\left(50\,mm, 20\,mm\right)$ and $\left(80\,mm, 40\,mm\right)$}.
The reported execution time is the average over $200$ continuous frames to remove any fluctuations caused by the different distributions of the athletes.
The creation of 2D billboards swamped the computation time of \cite{Sankoh2018Acmmm}.
It created 130 and 32 billboards for the volleyball and judo {sequences,} respectively.   
As for \cite{smolic20113d}, the noise filtering is the most time-consuming component that contributes to the production {and} accounted for around $90\%$
and $50\%$ for the volleyball and judo {sequences,} respectively.
The different {percentages} indicate that {the} execution time changes significantly as the resolution of {the} pre-defined volume increases.
Comparing these results, the proposed method {can be executed more quickly than the other methods}.
Moreover, {it demonstrated consistent performances for the three sequences}.
The average execution times {for each of the production processes} are shown in Table~\ref{tab:time-2}. 
These times indicate that the noise filtering and ROI extraction still consume the most time, especially for a sports {event taking place} in a large space. 



\begin{table}[t]
\caption{Execution time for each process in milliseconds.}
\label{tab:time-2}
\centering
{\footnotesize
\begin{tabular}{|l|c|c|c|}
\hline
 & 1st V & 2nd V & Judo   \\
\hline\hline
(B-1) Sparse visual hull &  0.54&  0.58&  0.06\\
\hline
(B-2) Noise filtering \& 3D ROI extraction &  21.49 &  19.75&  2.11\\
\hline
(B-3) Dense visual hull &  1.23&  1.13&  0.23 \\
\hline
(C) Polygonization &  3.90&  3.67 &  0.65 \\
\hline
(D-1) Computation of depth image & 5.93 &  5.52 &  2.99\\
\hline
(D-2) Visibility detection & 3.30 &  3.01 &  1.26\\
\hline
\end{tabular}
}
\vskip -4mm
\end{table}

\subsection{Discussion {of} parameters}

\begin{figure}[t]
\centering
\footnotesize
	\begin{minipage}[b]{0.85\linewidth}
 		\centering
 		\subfloat[{different voxel sizes for sparse visual hull}]
		{
 	 		\begin{overpic}[width=1\textwidth]
 	 			{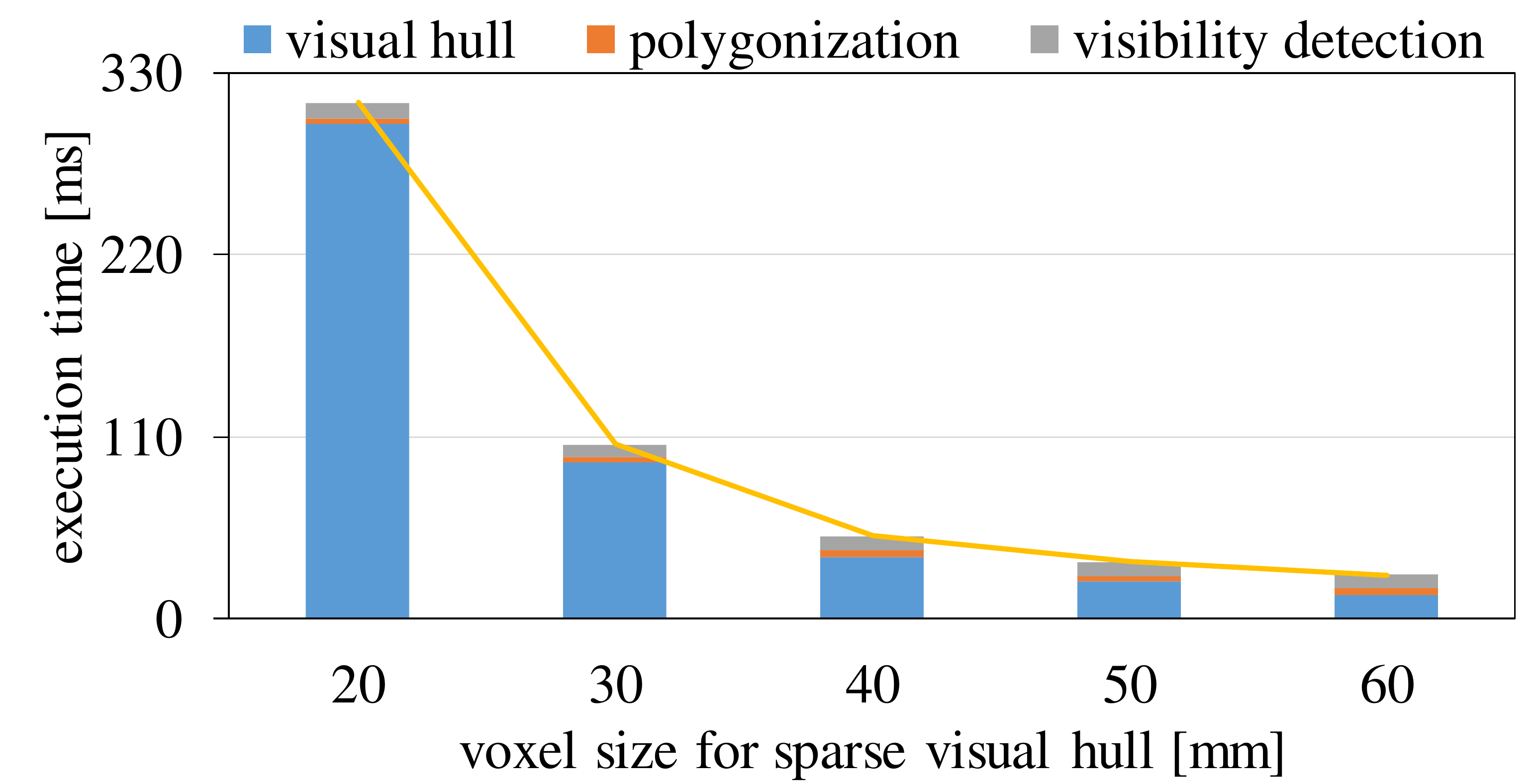}
 		\end{overpic}
 	 	}
	\end{minipage}
\vskip 2mm
	\begin{minipage}[b]{0.85\linewidth}
 		\centering
 		\subfloat[{different voxel sizes for dense visual hull}]
		{
 			\begin{overpic}[width=1\textwidth]
 	 			{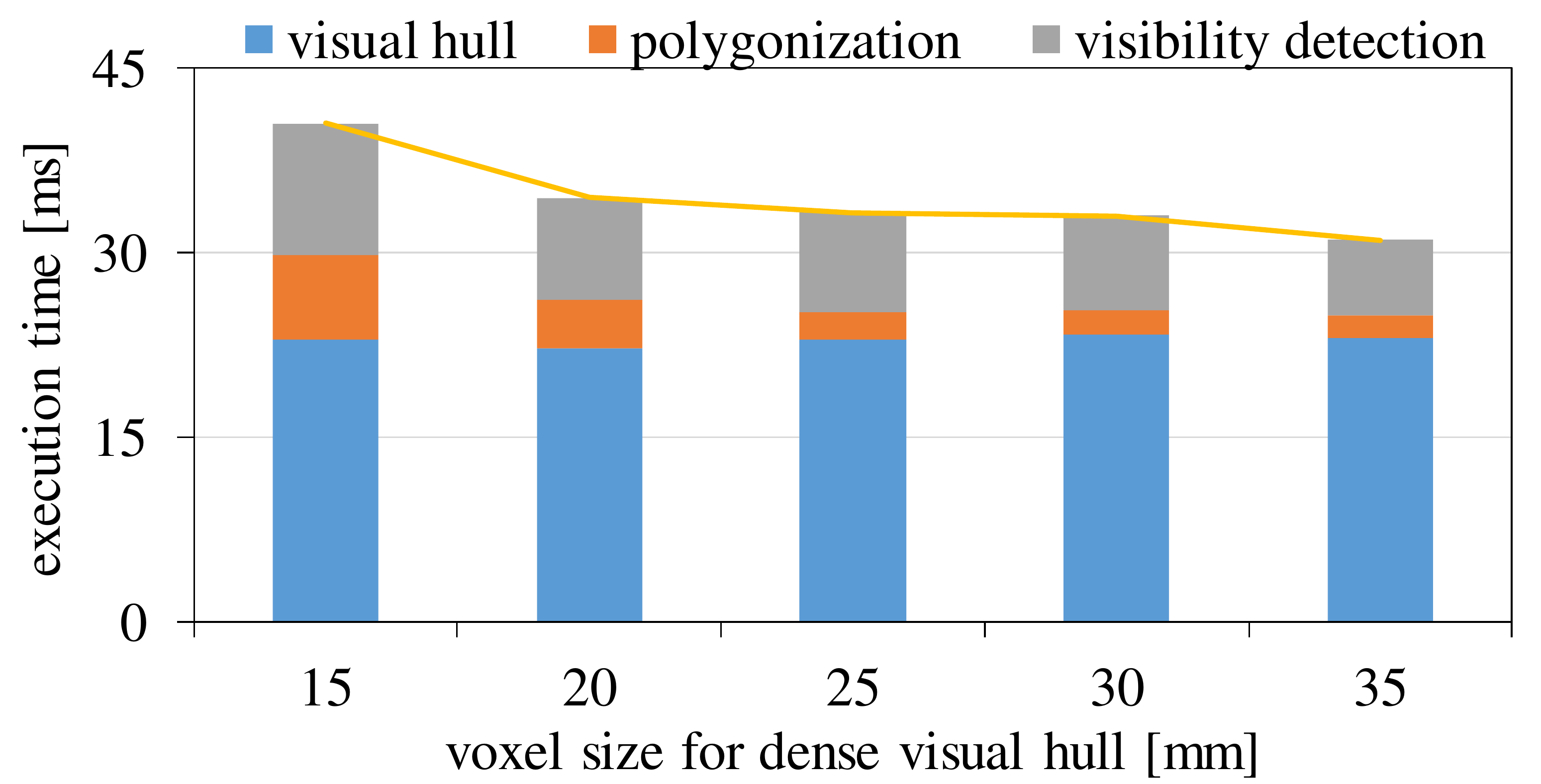}
 		\end{overpic}
 		}
	\end{minipage}	
\caption{Execution time versus different voxel sizes.}
\label{fig:execution-versus-parameters}
\vskip -4mm
\end{figure}

To clarify the effect of voxel sizes on execution time, we performed another two experiments with the first volleyball {sequence}.
In the first experiments, we conducted FVV production {while} varying the voxel size for sparse visual hull from $20~$mm to $60~$mm (stride is $10~$mm) {but} keeping the voxel size for dense visual hull {steady} at $20~$mm.
Fig.~\ref{fig:execution-versus-parameters} (a) shows the execution times.
{When the voxel size for both coarse and fine visual hull construction {is} $20~$mm, the {acceleration function} is disabled.
The time consumed {in} this case is around $310~$ms, which was the highest among the tests.
The time curve proves that the execution time has a positive association with the voxel size for sparse visual hull.}
It can be reduced by increasing the voxel size.
Fig.~\ref{fig:execution-versus-parameters} (b) presents the time consumed {in} the second experiments, in which we varied the voxel size for dense visual hull from $15~$mm to $35~$mm (stride is $5~$mm) {but kept} the voxel size for sparse visual hull {steady} at $50~$mm.
{The results} indicate that the number of voxels in dense visual hull construction does not {have a marked effect on production time}.

\section{CONCLUSIONS}

In this paper, we proposed a novel parallel approach to solve the fast FVV synthesis issue {that occurs in sports scenario}.
Our method first employs a {sparse-to-fine} volumetric visual hull construction strategy to reduce the time {needed for} 3D shape approximation and then applies an accurate isosurface extraction method to give a high-quality mesh representation to the 3D model.
Coupled with the occlusion map in each camera, the appearance of the scene at a virtual viewpoint is reproduced in a view-dependent manner.
We implemented the proposed method on a PC with a GPU and verified its performance with volleyball and judo {sequences}.
The experimental results show that our method outperforms existing approaches in terms of both visual quality and execution time.

\addtolength{\textheight}{-12cm}   





\bibliographystyle{IEEEbib}
\bibliography{mybibfile.bib}

\end{document}